\documentclass[lettersize,journal]{IEEEtran}
\usepackage{amsmath,amsfonts}
\usepackage{array}
\usepackage[caption=false,font=normalsize,labelfont=sf,textfont=sf]{subfig}
\usepackage{textcomp}
\usepackage{stfloats}
\usepackage{url}
\usepackage{verbatim}
\usepackage{graphicx}
\usepackage{cite}
\usepackage{multirow} % 用于跨行
\usepackage{pifont} % 导入pifont宏包
\usepackage{colortbl} % 用于单元格颜色
\usepackage{xcolor}
\usepackage{algorithm}
\usepackage{algpseudocode}
\usepackage{caption}

\usepackage{wasysym}   % For \CIRCLE
\usepackage{pdfpages}  % For PDF

\usepackage{hyperref}
\hypersetup{
	colorlinks=true,
	linkcolor=cyan,
	filecolor=blue,      
	urlcolor=black,
	citecolor=green,
}

\usepackage{etoolbox}
\patchcmd{\thebibliography}{\settowidth}
    {\setlength{\itemsep}{0pt plus 0.3ex}\settowidth}{}{}
\def\IEEEbibitemsep{0pt plus .5pt}
\let\oldthebibliography\thebibliography
\renewcommand{\thebibliography}[1]{%
  \oldthebibliography{#1}%
  \setlength{\itemsep}{\IEEEbibitemsep}%
  \scriptsize % 调整文献字体大小
}

\usepackage{tikz}
\newcommand{\circEmpty}{%
\tikz[baseline=-0.6ex] \draw[line width=0.4pt] (0,0) circle (0.8ex);
}

\newcommand{\circHalf}{%
\tikz[baseline=-0.6ex]{
    \draw[line width=0.4pt] (0,0) circle (0.8ex);
    \fill (0,0) -- +(270:0.8ex) arc (270:90:0.8ex) -- cycle;
}
}

\newcommand{\circFull}{%
\tikz[baseline=-0.6ex] \fill (0,0) circle (0.8ex);
}
\newcommand{\circExcellent}{%
\tikz[baseline=-0.6ex]{
    \fill (0,0) circle (0.55ex);                    % solid core
    \draw[line width=0.5pt] (0,0) circle (0.85ex);  % outer ring
}
}

\newcommand{\Low}{\circEmpty}
\newcommand{\Medium}{\circHalf}
\newcommand{\High}{\circFull}
\newcommand{\Excellent}{\circExcellent}  % 双实心表示最强

\newcommand{\crbx}[1]{\scalebox{0.75}{\fcolorbox{#1}{#1}{\rule{0pt}{1.5ex}\rule{1.5ex}{0pt}}}}

\definecolor{ts1m_purple}{RGB}{138,83,192}
\definecolor{ts1m_green}{RGB}{78,167,46}
\definecolor{ts1m_blue}{RGB}{52,204,204}
\definecolor{ts1m_gray}{RGB}{165,165,165}
\definecolor{ts1m_red}{RGB}{192,0,0}
\definecolor{ts1m_yellow}{RGB}{248,190,50}

\definecolor{best}{HTML}{FFF2CC}    % \cellcolor[HTML]{FFF2CC}
\definecolor{second}{HTML}{F1E9DF}  % \cellcolor[HTML]{F1E9DF}

\newcommand{\best}{\crbx{best}}
\newcommand{\second}{\crbx{second}}

\begin{document}

% \title{Diagnostic Benchmark for Traffic Sign Recognition: TS-1M, Challenge Suites, and Field Validation}
\title{Traffic Sign Recognition in Autonomous Driving: Dataset, Benchmark, and Field Experiment}

% \author{Author Names Omitted for Anonymous Review}

\author{
Guoyang~Zhao$^*$,
Weiqing~Qi$^*$,
Kai~Zhang,
Chenguang~Zhang,
Zeying~Gong,
Zhihai~Bi,
Kai~Chen,
Benshan~Ma,
Ming~Liu,~\IEEEmembership{Fellow,~IEEE},
and Jun~Ma,~\IEEEmembership{Senior~Member,~IEEE}
\IEEEcompsocitemizethanks{
\IEEEcompsocthanksitem Guoyang~Zhao, Weiqing~Qi, Kai~Zhang, Zeying~Gong, Zhihai~Bi, Kai~Chen, and Benshan~Ma are with The Hong Kong University of Science and Technology (Guangzhou), Guangzhou 511453, China (e-mail: gzhao492@connect.hkust-gz.edu.cn).
\IEEEcompsocthanksitem Chenguang~Zhang is with Lingnan University, Hong Kong SAR, China (e-mail: qwe934063437@gmail.com).
% \IEEEcompsocthanksitem Ming Liu is with the Research \& Development Institute of Northwestern Polytechnical University, Shenzhen 518063, China (e-mail: liu.ming.prc@gmail.com).
\IEEEcompsocthanksitem Ming Liu is with Shenzhen Unity Drive Innovation Technology Co., Ltd.,
        Shenzhen 518063, China (e-mail: liu.ming.prc@gmail.com).
\IEEEcompsocthanksitem Jun Ma is with The Hong Kong University of Science and Technology (Guangzhou), Guangzhou 511453, China, and also with The Hong Kong University of Science and Technology, Hong Kong SAR, China (e-mail: jun.ma@ust.hk). 
\IEEEcompsocthanksitem * indicates equal contribution.
}
}

\IEEEaftertitletext{%
\vspace{-3.0em} % 缩小图与作者的距离
\noindent\begin{center}
\includegraphics[width=\linewidth]{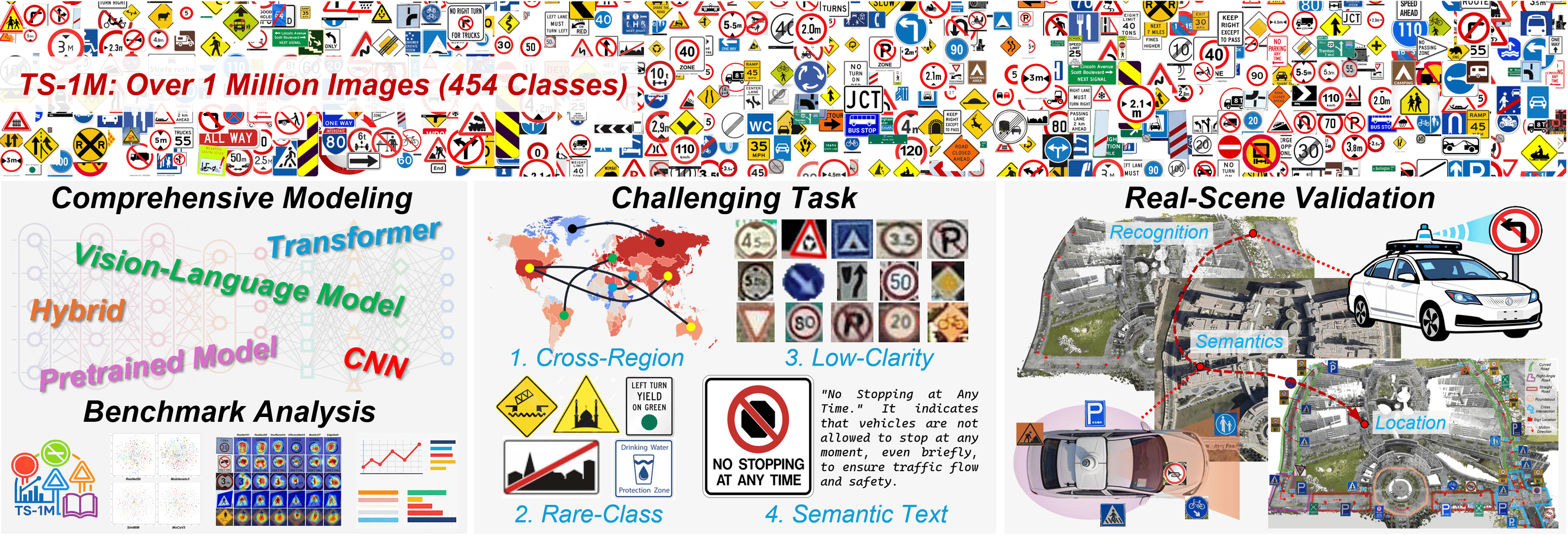}
\captionof{figure}{
\textbf{A diagnostic benchmark for traffic sign recognition.}
The dataset contains over one million images across 454 categories and supports comprehensive benchmarking of diverse model families under four key challenge suites.
An on-road case study illustrates the integration of recognition, rule generation, and spatial localization for semantic-map construction.
}
\label{cover-figure}
\end{center}
% \vspace{0.5em} % 视需要微调与摘要间距
% \vspace{-2pt}
}

\maketitle

\begin{abstract}
Traffic Sign Recognition (TSR) is a core perception capability for autonomous driving, where robustness to cross-region variation, long-tailed categories, and semantic ambiguity is essential for reliable real-world deployment.
Despite steady progress in recognition accuracy, existing traffic sign datasets and benchmarks offer limited diagnostic insight into how different modeling paradigms behave under these practical challenges.
We present TS-1M, a large-scale and globally diverse traffic sign dataset comprising over one million image samples across 454 standardized categories, together with a diagnostic benchmark designed to analyze model capability boundaries.
Beyond standard train-test evaluation, we provide a suite of challenge-oriented settings, including cross-region and cross-dataset transfer, rare-class identification, low-clarity robustness, and recognition with semantic descriptions, enabling systematic and fine-grained assessment of modern TSR models.
Using TS-1M, we conduct a unified benchmark across three representative learning paradigms: classical supervised models, self-supervised pretrained models, and multimodal vision-language models (VLMs).
Our analysis reveals model-dependent performance under target-held-out transfer and rare-category recognition, with CLIP showing strong performance across both settings.
Finally, an on-road semantic-mapping case study illustrates how TS-1M-based recognition, generated rule texts, and spatial localization can be integrated into map annotations.
Overall, TS-1M establishes a reference-level diagnostic benchmark for TSR and provides principled insights into robust and semantic-aware traffic sign perception. 
% Project page: \href{https://anonymous.4open.science/w/ts1m-anonymous-review-20C8/}{\textcolor[RGB]{0,153,255}{anonymous.4open.science/w/ts1m-anonymous-review-20C8}}.
\textit{Project page:} 
{\hypersetup{urlcolor=blue}\url{https://guoyangzhao.github.io/projects/ts1m}}.
% Project resources: \href{https://anonymous.4open.science/w/ts1m-anonymous-review-20C8/}{\textcolor[RGB]{0,153,255}{anonymous page}}.
\end{abstract}

\begin{IEEEkeywords}
Traffic sign recognition, deep learning, dataset, benchmark analysis, autonomous driving.
\end{IEEEkeywords}

\section{Introduction}
\label{sec:intro}

Traffic Sign Recognition (TSR) plays a crucial role in autonomous driving by enabling vehicles to interpret regulatory, warning, and guidance signs for safe navigation and scene understanding \cite{sakaridis2025acdc, aghdam2017practical}.
Early TSR methods relied on handcrafted visual features and parameter-based classifiers \cite{mogelmose2015detection, guo2023grtr}, which were sensitive to environmental variation and required manual tuning.
Recent advances in deep learning, particularly Convolutional Neural Networks (CNNs), have substantially improved TSR performance by learning discriminative representations from large-scale data \cite{he2016deep}.
By fine-tuning models pretrained on ImageNet \cite{krizhevsky2012imagenet}, prior studies have achieved strong accuracy on benchmark datasets \cite{almutairy2019arts, wang2024traffic, wang2024dk}.
However, high accuracy on benchmark datasets does not necessarily imply robust generalization in real-world driving environments~\cite{zhang2024coda}.

A key limitation arises from existing traffic sign datasets, which are typically collected within restricted geographic regions and exhibit limited scale, incomplete category coverage, and inconsistent labeling standards \cite{shakhuro2016russian, tabernik2019deep, almutairy2019arts}. 
As a result, models trained on these datasets often suffer significant performance degradation when evaluated across regions or under diverse environmental conditions, even for nominally identical sign categories \cite{zhao2024tsclip}. 
Moreover, differences in category definitions and annotation protocols complicate the integration of multiple datasets into a unified benchmark~\cite{singh2022road,sanchez2025cola}.
In addition, many datasets provide only symbolic labels without explicit semantic descriptions \cite{houben2013detection, zhu2016traffic, shakhuro2016russian, almutairy2019arts}, which limits their applicability to driving scenarios that require semantic understanding of traffic rules.

Recent robotic datasets, including CODa for urban perception~\cite{zhang2024coda}, eAP for event-enhanced autonomous perception~\cite{li2026eap}, and ROVER for multiseason visual SLAM~\cite{schmidt2025rover}, highlight the value of diverse data and task-specific benchmarks.
Motivated by this trend and the limitations of existing traffic sign datasets, we introduce TS-1M, a large-scale dataset designed to support diagnostic evaluation and foundational research in traffic sign recognition.
TS-1M is constructed by consolidating and harmonizing images from mainstream traffic sign datasets \cite{zhu2016traffic, houben2013detection, safavi2025persian, timofte2014multi, shakhuro2016russian, tusher2023comparative, ertler2020mapillary, tabernik2019deep, larsson2011using, almutairy2019arts, mogelmose2012vision} and diverse Internet sources \cite{erdem2023trafficsignturkey, nlpr_chinese_traffic_signs, jodh2023indiantrafficsigns, maitam2023vietnamesetrafficsigns, buqi2023trafficsignclassification, joshi2023canadianroadsigns, kasia2023polishtrafficsigns, sachsene2023carlatrafficsigns, simonin2023trafficsigndatabase, cakrulgaming2023indonesiatrafficsigns, caballa2023britishcolumbiatrafficsigns}. 
After extensive cleaning and standardization according to regional classification standards, TS-1M comprises over one million images across 454 traffic sign categories, and currently represents the most geographically diverse and category-rich traffic sign dataset available, enabling large-scale and cross-regional benchmarking.
TS-1M also provides a set of challenge-oriented subsets that explicitly support diagnostic analysis, including cross-region recognition, rare-category difficulty, low-clarity robustness, and recognition with semantic descriptions.

In addition to dataset limitations, practical TSR systems face several emerging challenges as application scenarios continue to expand.
\textbf{(a) Cross-region generalization:} Models trained in one geographic region often fail to generalize to others due to variations in sign appearance, design standards, and environmental context \cite{zhao2024tsclip}.
\textbf{(b) Long-tailed and rare-category recognition:} Many traffic sign categories are inherently underrepresented \cite{gao2022long, wang2024more}, making reliable recognition of rare signs difficult for existing approaches.
\textbf{(c) Robustness to low-clarity observations:} Traffic signs frequently appear blurred, occluded, or degraded due to distance, motion, and environmental interference, posing significant challenges to model robustness \cite{ahmed2021dfr, gray2023glare}.
\textbf{(d) Recognition with semantic descriptions:} Traffic signs convey information relevant to rule-compliant motion planning~\cite{lin2025trafficrepair}; we examine how scene and rule descriptions affect traffic sign classification~\cite{yang2024traffic}.
Despite their practical importance, these challenges have not been systematically analyzed within a unified and diagnostic evaluation framework.

In this paper, we present TS-1M, a large-scale and globally diverse traffic sign dataset, together with a unified diagnostic benchmark for systematic TSR evaluation under realistic conditions, as illustrated in Fig.~\ref{cover-figure}. 
The benchmark spans three representative learning paradigms, including classical supervised models, self-supervised pretrained models, and recent multimodal VLMs, and evaluates them across four challenge-oriented settings: cross-region and cross-dataset generalization, rare-category recognition, low-clarity robustness, and recognition with semantic descriptions. 
Through comprehensive empirical analysis, we characterize paradigm-dependent performance trends and capability boundaries in TSR, providing a more diagnostic understanding of the strengths and limitations of existing approaches. 
An on-road case study further illustrates how TS-1M-based recognition can support semantic-map construction by associating sign categories and generated rule texts with spatial locations.
The main contributions of this paper are summarized as follows:
\begin{enumerate}
    \item We construct TS-1M, a large-scale and globally diverse traffic sign dataset with over one million images across 454 standardized categories, providing a unified data foundation for diagnostic TSR benchmarking.
    \item We identify and formalize key challenges in large-scale TSR from a diagnostic perspective, including cross-region generalization, long-tailed recognition, low-clarity robustness, and recognition with semantic descriptions.
    \item We establish a unified diagnostic benchmark on TS-1M to evaluate representative TSR modeling paradigms under realistic and challenge-oriented settings. Transfer evaluation uses separate training runs that exclude each target dataset and, for geographically localized targets, its corresponding region.
    \item Through extensive empirical analysis and real-scene experiments, we characterize paradigm-dependent performance trends and capability boundaries in TSR, highlighting the emerging advantages of VLMs in robust and semantic-aware recognition.
\end{enumerate}

\section{Related Works}
\label{sec:related}

\subsection{Traffic Sign Recognition}
Early TSR systems predominantly relied on handcrafted visual cues (e.g., color, shape, and edge patterns) combined with parametric classifiers~\cite{mogelmose2015detection}. Although effective in controlled settings, such pipelines require substantial manual feature engineering and domain expertise, and typically exhibit limited robustness under illumination changes, viewpoint variations, and cluttered driving scenes~\cite{guo2023grtr}. 

Deep learning has substantially advanced TSR by enabling end-to-end feature learning with CNNs, autoencoders, and multi-scale architectures~\cite{wang2024more}. Beyond improving accuracy, recent studies have explored robustness-oriented designs, including two-stage or fusion-based pipelines with refined post-processing for low-resolution inputs~\cite{li2024toward,wang2020traffic}, learning strategies for long-tailed distributions and category imbalance~\cite{wang2024more}, attention-based feature refinement for geometric deformation and viewpoint skew~\cite{an2024road}, defenses against adversarial perturbations~\cite{sarwatt2024adapting}, and temporal modeling using multi-frame cues in video streams~\cite{yu2022traffic}. In parallel, incorporating scene semantics and spatial constraints improves structured recognition and localization in complex environments~\cite{min2022traffic}. 

Despite this progress, several fundamental challenges remain in real-world TSR and are closely related to our benchmark design. 
\textbf{Cross-region and cross-scenario generalization} remains difficult due to variations in sign appearance, language, and regional design standards; representative efforts include domain generalization and knowledge transfer~\cite{abdi2024seqnet}, multi-task detection-classification learning under region-specific styles~\cite{uikey2024indian}, and multimodal adaptation via CLIP-style fine-tuning to mitigate cross-lingual and cross-cultural shifts~\cite{zhao2024tsclip}. 
\textbf{Rare-class and few-shot recognition} is hindered by long-tailed distributions and limited samples; existing solutions enhance feature fusion and sampling or optimization strategies to improve learning under data scarcity~\cite{wang2023vehicle,gao2022long,wang2024dk}. 
\textbf{Robust recognition under low-clarity conditions} (e.g., motion blur, adverse weather, low resolution, and glare) remains a persistent challenge. Prior work explores image enhancement and photometric filtering~\cite{zhang2024robust}, analyzes illumination as a practical threat factor~\cite{hsiao2024natural}, introduces datasets for glare-induced degradation~\cite{gray2023glare}, and combines enhancement with multi-stage modeling or semi-supervised learning to mitigate quality degradation~\cite{ahmed2021dfr,chen2024semi}. 
Finally, \textbf{semantic text understanding} extends TSR beyond appearance-based classification by leveraging textual content and rule-level semantics. Recent efforts include multi-task learning for generating natural language descriptions~\cite{yang2024traffic}, efficient text localization and correction in multi-scenario settings~\cite{han2024real}, and zero-shot TSR guided by design standards and prior knowledge to improve adaptability to unseen categories~\cite{cao2022sustainable}. 

Overall, existing TSR studies largely focus on individual challenges. A broader benchmark is needed to examine regional transfer, rare-class recognition, low-clarity robustness, and recognition with semantic descriptions within a unified evaluation framework.

\subsection{Traffic Sign Dataset}
In recent years, both academia and industry have developed numerous traffic sign datasets across different countries and regions, providing essential support for TSR research and evaluation. Table~\ref{Table-Dataset} summarizes their key attributes. Widely used benchmarks include STS \cite{larsson2011using}, LISA \cite{mogelmose2012vision}, GTSRB \cite{houben2013detection}, BelgiumTS \cite{timofte2014multi}, TT100K \cite{zhu2016traffic}, RTSD \cite{shakhuro2016russian}, DFG \cite{tabernik2019deep}, ARTS \cite{almutairy2019arts}, MTSD \cite{ertler2020mapillary}, BanglaTS \cite{tusher2023comparative}, and PTSD \cite{safavi2025persian}. These datasets differ in geographic coverage, category scale, image quality, and annotation standards, forming the foundation of TSR systems. Additionally, specialized datasets targeting challenges such as adverse lighting or fine-grained recognition facilitate scenario-specific analysis.

From a geographic and taxonomic perspective, existing datasets exhibit complementary strengths but also clear limitations. Region-specific benchmarks such as STS and DFG provide high-quality annotations but are limited in scale and diversity~\cite{larsson2011using,tabernik2019deep}. North-America-centric datasets (e.g., LISA and ARTS) are effective for U.S.-style environments but lack multilingual and cross-standard coverage~\cite{mogelmose2012vision,almutairy2019arts}. Classical benchmarks such as GTSRB offer large sample sizes but cover relatively narrow taxonomies, restricting fine-grained evaluation~\cite{houben2013detection}. Other datasets (e.g., BelgiumTS, TT100K, and RTSD) capture regional characteristics in Europe, China, and Russia, yet often suffer from limited scale, regional bias, or class imbalance, hindering cross-region transfer~\cite{timofte2014multi,zhu2016traffic,shakhuro2016russian}. More recent datasets from underrepresented regions (e.g., BanglaTS and PTSD) highlight language- and culture-specific patterns but remain insufficient in scale and coverage~\cite{tusher2023comparative,safavi2025persian}. MTSD provides broader cross-country diversity; however, annotation inconsistency, redundant categories, and limited per-class samples still constrain its use as a unified large-scale benchmark~\cite{ertler2020mapillary}.

TSCLIP~\cite{zhao2024tsclip} introduced CRTS, a cross-region benchmark comprising over 190,000 images from ten source datasets, harmonized into 46 common categories. Its CLIP adaptation incorporates scene and traffic-rule descriptions. CRTS supports recognition of shared sign categories across regions, while its 46-category scope leaves many fine-grained and region-specific signs outside evaluation.

TS-1M broadens this scope through worldwide collection from public datasets, dataset platforms, and diverse Internet sources~\cite{nlpr_chinese_traffic_signs,jodh2023indiantrafficsigns,maitam2023vietnamesetrafficsigns,buqi2023trafficsignclassification,erdem2023trafficsignturkey,joshi2023canadianroadsigns,kasia2023polishtrafficsigns,sachsene2023carlatrafficsigns,simonin2023trafficsigndatabase,cakrulgaming2023indonesiatrafficsigns,caballa2023britishcolumbiatrafficsigns}. Label harmonization yields 454 categories and 1,267,558 image samples, including offline augmentation. Beyond cross-region recognition, TS-1M provides rare-class, low-clarity, and semantic-description evaluations under a shared benchmark protocol for supervised, self-supervised, and vision-language models.

\begin{table}[t]
\renewcommand\arraystretch{1.3}
\caption{\textbf{Comparison of Mainstream TSR Datasets.}}
\vspace{-5pt}
\centering
\setlength{\tabcolsep}{1.4mm} % Adjust column separation
\footnotesize
\begin{tabular}{ccccccc}
\hline
\rowcolor{ts1m_purple!7} \textbf{No.} & \textbf{Name} & \textbf{Region} & \begin{tabular}[c]{@{}c@{}}\textbf{Class}\\\textbf{Num.}\end{tabular} & \begin{tabular}[c]{@{}c@{}}\textbf{Semantic}\\\textbf{Name}\end{tabular} & \textbf{Examples} & \textbf{Year} \\
\hline
1 & STS \cite{larsson2011using} & Sverige & 20 & \textcolor[HTML]{228B22}{\ding{51}} & 5582 & 2011 \\
\rowcolor{gray!9} 2 & LISA \cite{mogelmose2012vision} & America & 47 & \textcolor[HTML]{228B22}{\ding{51}} & 7855 & 2012 \\
3 & GTSRB \cite{houben2013detection} & Germany & 43 & \textcolor[HTML]{B22222}{\ding{55}} & 51839 & 2013 \\
\rowcolor{gray!9} 4 & BelgiumTS \cite{timofte2014multi} & Belgium & 62 & \textcolor[HTML]{B22222}{\ding{55}} & 7125 & 2014 \\
5 & TT100K \cite{zhu2016traffic} & China & 127 & \textcolor[HTML]{B22222}{\ding{55}} & 30000 & 2016 \\
\rowcolor{gray!9} 6 & RTSD \cite{shakhuro2016russian} & Russia & 106 & \textcolor[HTML]{B22222}{\ding{55}} & 105509 & 2016 \\
7 & BrazilianTS \cite{simonin2023trafficsigndatabase} & Brazilian & 33 & \textcolor[HTML]{228B22}{\ding{51}} & 3351 & 2018 \\
\rowcolor{gray!9} 8 & DFG \cite{tabernik2019deep} & Slovenia & 200 & \textcolor[HTML]{B22222}{\ding{55}} & 13239 & 2019 \\
9 & ARTS \cite{almutairy2019arts} & America & 175 & \textcolor[HTML]{B22222}{\ding{55}} & 36187 & 2019 \\
\rowcolor{gray!9} 10 & CanadaTS \cite{joshi2023canadianroadsigns} & Canada & 39 & \textcolor[HTML]{228B22}{\ding{51}} & 204 & 2019 \\
11 & PolishTS \cite{kasia2023polishtrafficsigns} & Polish & 92 & \textcolor[HTML]{B22222}{\ding{55}} & 21044 & 2020 \\
\rowcolor{gray!9} 12 & TurkeyTS \cite{erdem2023trafficsignturkey} & Turkey & 91 & \textcolor[HTML]{B22222}{\ding{55}} & 21249 & 2020 \\
13 & MTSD \cite{ertler2020mapillary} & World & 313 & \textcolor[HTML]{228B22}{\ding{51}} & 206388 & 2020 \\
\rowcolor{gray!9} 14 & TSCR \cite{buqi2023trafficsignclassification} & China & 10 & \textcolor[HTML]{B22222}{\ding{55}} & 6348 & 2021 \\
15 & CarlaTS \cite{sachsene2023carlatrafficsigns} & Germany & 8 & \textcolor[HTML]{228B22}{\ding{51}} & 2977 & 2021 \\
\rowcolor{gray!9} 16 & IndonesiaTS \cite{cakrulgaming2023indonesiatrafficsigns} & Indonesia & 21 & \textcolor[HTML]{228B22}{\ding{51}} & 1800 & 2021 \\
17 & BritishTS \cite{caballa2023britishcolumbiatrafficsigns} & British & 8 & \textcolor[HTML]{228B22}{\ding{51}} & 2217 & 2021 \\
\rowcolor{gray!9} 18 & IndiaTS \cite{jodh2023indiantrafficsigns} & India & 85 & \textcolor[HTML]{228B22}{\ding{51}} & 5726 & 2022 \\
19 & CTSRD \cite{nlpr_chinese_traffic_signs} & China & 58 & \textcolor[HTML]{B22222}{\ding{55}} & 5998 & 2023 \\
\rowcolor{gray!9} 20 & VNTS \cite{maitam2023vietnamesetrafficsigns} & Vietnam & 51 & \textcolor[HTML]{B22222}{\ding{55}} & 7784 & 2023 \\
21 & BanglaTS \cite{tusher2023comparative} & Bangla & 13 & \textcolor[HTML]{228B22}{\ding{51}} & 8386 & 2023 \\
\rowcolor{gray!9} 22 & PTSD \cite{safavi2025persian} & Iran & 43 & \textcolor[HTML]{228B22}{\ding{51}} & 16421 & 2025 \\
\hline
- & \textbf{TS-1M (Train)} & World & 454 & \textcolor[HTML]{228B22}{\ding{51}} & 1033947 & 2026 \\
\rowcolor{gray!9} - & \textbf{TS-1M (Test)} & World & 454 & \textcolor[HTML]{228B22}{\ding{51}} & 233611 & 2026 \\
\hline
\end{tabular}
\label{Table-Dataset}
\vspace{-10pt}
\end{table}

\begin{figure*}[t!]
    \centering
    \includegraphics[width=.99\textwidth]{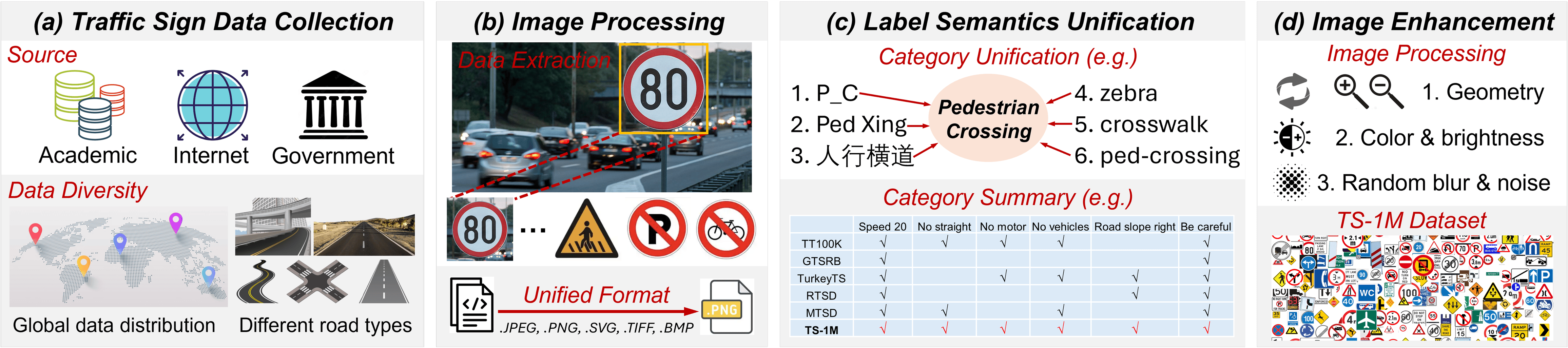}
    \vspace{-2pt}
    \caption{\textbf{Overview of the TS-1M dataset construction process.} 
The pipeline covers multi-source data collection, unified image preprocessing, semantic label normalization across heterogeneous datasets, and data augmentation to support large-scale, consistent, and robust benchmarking for TSR.}
    \label{dataset-build}
    \vspace{-15pt}
\end{figure*}

\subsection{Deep Learning Benchmark Model}

To systematically evaluate different learning paradigms under realistic TSR challenges, we benchmark representative architectures from three families: (i) classical supervised visual backbones (CNNs and Transformers), (ii) self-supervised pretrained models, and (iii) multimodal VLMs. Leveraging the large-scale, multi-region, and multi-condition nature of TS-1M, this benchmark characterizes the capability boundaries of each family under cross-region shift, long-tailed distributions, low-clarity inputs, and recognition with semantic descriptions. To ensure fair and practical comparison, we select widely adopted, open-source, and fine-tuning-friendly models spanning a broad accuracy–efficiency spectrum.

\textbf{Classical Deep Learning Architectures}. 
Supervised CNN and Transformer backbones remain the standard for TSR due to strong in-domain discriminative ability and mature deployment pipelines. We include representative CNN families from high-capacity to lightweight variants: ResNet~\cite{he2016deep} as a residual baseline, ResNeXt~\cite{xie2017aggregated} with aggregated transformations, and ConvNeXt~\cite{liu2022convnet} as a modernized convolutional architecture competitive with Transformers. For real-time and embedded scenarios, MobileNet~\cite{howard2019searching} and ShuffleNet~\cite{ma2018shufflenet} provide efficient alternatives. 
Transformer-based architectures are also considered for modeling long-range dependencies and global context. Swin Transformer~\cite{liu2021swin} introduces hierarchical windowed attention for multi-scale representation, while MobileViT~\cite{mehta2021mobilevit} and EdgeNeXt~\cite{maaz2022edgenext} explore hybrid designs balancing accuracy and efficiency on edge devices. 
Together, these models define the performance ceiling and deployment trade-offs of purely visual approaches on TS-1M.

\textbf{Self-Supervised Pretraining Architectures}. 
Self-supervised learning (SSL) enables transferable representations from large-scale unlabeled data, particularly relevant for TSR under long-tailed distributions and domain shifts. We include the contrastive methods SimCLR~\cite{chen2020simple} and MoCoV3~\cite{chen2021empirical}, and the self-distillation method DINO~\cite{caron2021emerging}.
We further consider masked image modeling approaches, including MAE~\cite{he2022masked} and SimMIM~\cite{xie2022simmim}, which learn global structure through reconstruction-based objectives with different efficiency–accuracy trade-offs. 
This setup evaluates whether SSL pretraining improves robustness across regions, rare categories, and degraded conditions.

\textbf{Multimodal Vision-Language Architectures}. 
VLMs jointly learn from images and text and exhibit strong open-vocabulary and cross-domain generalization, making them suitable for TSR with symbolic and textual semantics. We include CLIP~\cite{radford2021learning}, which aligns image–text embeddings via large-scale contrastive pretraining and supports prompt-based zero-shot recognition. 
Beyond contrastive models, the BLIP family~\cite{li2022blip} combines alignment with generative modeling, while BLIP2~\cite{li2023blip} introduces a Querying Transformer to interface with large language models for enhanced semantic grounding. LLaVA~\cite{liu2023visual} further extends this paradigm by enabling instruction-following multimodal reasoning, particularly for semantic-text understanding. 
We focus on mature, open-source, and fine-tuning-friendly models to ensure reproducible and meaningful evaluation. 
Overall, VLMs form a key component of our benchmark to quantify how multimodal semantic priors improve cross-region robustness and semantic-aware recognition on TS-1M.

\section{TS-1M Dataset}
\label{sec:dataset}

We construct a large-scale and unified traffic sign dataset, termed TS-1M, by integrating and standardizing images from published traffic sign datasets and additional public Internet sources. Table~\ref{Table-Dataset} compares representative datasets with TS-1M in geographic coverage, category coverage, and scale. TS-1M contains over one million images spanning 454 traffic sign categories. As illustrated in Fig.~\ref{dataset-build}, the dataset construction pipeline consists of three main stages: data collection, data preprocessing, and dataset release and statistics.

\subsection{Data Collection}

TS-1M combines images from published academic traffic sign datasets, additional collections hosted on public dataset platforms, governmental open-data platforms, and other publicly accessible Internet sources. These additional sources extend the collection beyond the representative datasets compared in Table~\ref{Table-Dataset}. Category-level manual review is used to harmonize labels across sources; its coverage and scope are specified in the data processing description below.

From a geographic perspective, TS-1M covers traffic sign images from a wide range of countries and regions worldwide, including Europe, Asia, North and South America, Africa, and the Middle East. Importantly, the dataset incorporates signs featuring local languages and region-specific designs, which is essential for evaluating cross-region generalization and multilingual robustness in real-world TSR.

In terms of scene diversity, the dataset includes images captured under a broad spectrum of real-world conditions, such as varying weather (e.g., sunny, rainy, and foggy), illumination (e.g., daylight, nighttime, and backlight), and driving environments (e.g., urban roads, rural areas, and highways). This diversity reflects realistic deployment scenarios and supports the development of models with strong robustness under complex and dynamic conditions.

Regarding category coverage, TS-1M consolidates major traffic sign types from existing datasets and reorganizes them into a unified, standardized, and hierarchical taxonomy. Through systematic reconciliation of heterogeneous labeling conventions, the final dataset comprises 454 fine-grained categories, forming one of the most comprehensive and diverse traffic sign category systems available to date.

To ensure long-term extensibility, TS-1M is built on a unified annotation schema that resolves semantic inconsistencies across heterogeneous sources and aligns disparate label definitions into a consistent representation. This design facilitates the integration of new datasets and supports the continuous expansion of sign categories and regional coverage, making TS-1M adaptable to evolving research demands in TSR.

\begin{figure*}[t!]
    \centering
    \includegraphics[width=.99\textwidth]{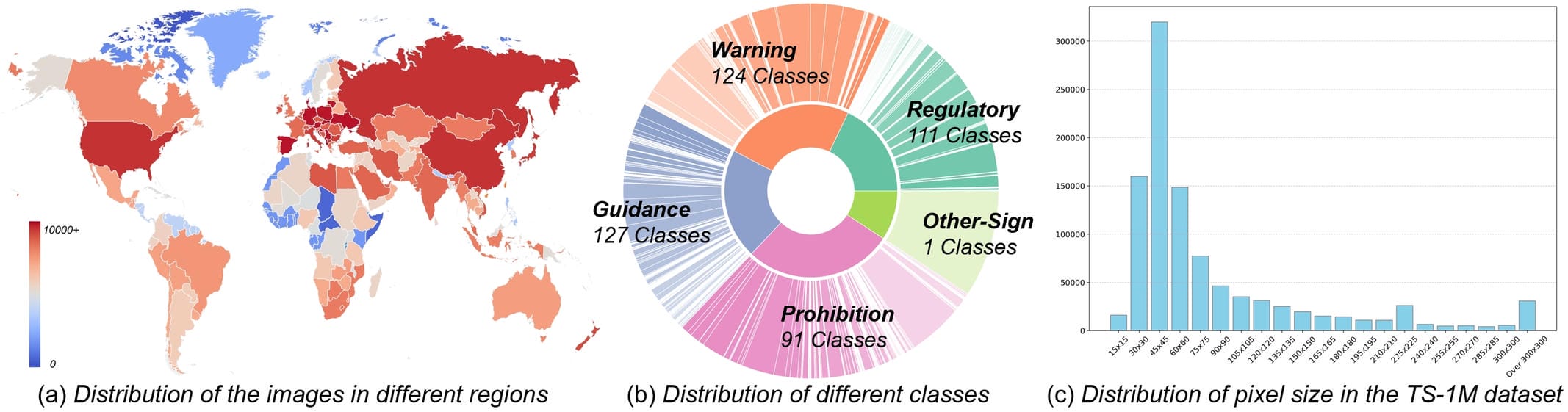}
    \vspace{-2pt}
    % \caption{\textbf{Overall of the TS-1M dataset.} xxx.}
    \caption{\textbf{Statistical overview of the TS-1M dataset.} 
The statistics summarize the spatial geographic distribution, category-wise sample distribution, and pixel-resolution distribution of images in TS-1M.}
    \label{dataset-statistics}
    \vspace{-15pt}
\end{figure*}

\subsection{Data Processing}

After data collection, we apply a comprehensive cleaning and preprocessing pipeline to ensure standardization, consistency, and high data quality across TS-1M, providing a reliable foundation for subsequent model training and evaluation.

First, all images are converted into standardized formats compatible with mainstream deep learning frameworks. Traffic sign regions are spatially normalized by cropping the sign areas according to the available annotations from the source datasets, reducing background interference and enhancing the visual saliency of target signs. This preprocessing step focuses on improving input consistency rather than introducing additional supervision.

Category definitions are harmonized with reference to national and international traffic sign standards~\cite{fhwa2023mutcd,unece2007roadsigns} and prior dataset taxonomies~\cite{houben2013detection,zhu2016traffic,ertler2020mapillary}. Source categories are aligned through joint inspection of sign graphics and traffic meanings. Semantically equivalent categories are merged, while differences in numerical values, directions, applicable vehicle types, or traffic functions are retained. All 454 retained categories are manually reviewed (100\% category-level coverage). 

To enhance robustness under diverse real-world conditions, we further apply an offline data augmentation strategy to expand the entire dataset by approximately 30\%. The augmented samples are generated using three commonly adopted augmentation techniques. Specifically, 40\% of the augmented data involve geometric transformations, including rotation, flipping, scaling, and translation. These transformations are constrained by class-specific semantics to preserve the sign category, including directional, textual, and numerical meaning. Another 40\% apply photometric adjustments such as brightness, contrast, and saturation changes to account for illumination variability. The remaining 20\% introduce blur and noise perturbations to mimic motion blur, sensor noise, and low-quality imaging conditions. This augmentation effectively enhances the generalization capability of models trained on TS-1M.

After data cleaning, label normalization, and augmentation, TS-1M adopts a fixed training and testing split at the original-image group level. Each original image and all of its augmented variants are assigned to the same partition, preventing overlap of an original image and its derived samples across training and testing. A common train/test split ratio is applied within each category, preserving the class-frequency distribution across the two partitions.

\subsection{Release \& Statistics}

\subsubsection{\textbf{TS-1M Core Dataset}}
The TS-1M core dataset consists of real-world traffic sign images collected from a wide range of countries and regions, covering diverse geographic areas, languages, and traffic regulations. The taxonomy comprises four principal functional groups---\emph{Restriction} (111 categories), \emph{Prohibition} (91), \emph{Warning} (124), and \emph{Guidance} (127)---plus one \emph{Other} category, giving 454 categories in total. This unified taxonomy reconciles heterogeneous labeling conventions across source datasets and provides a globally consistent classification system for traffic signs.

In terms of scale, the TS-1M core dataset contains a total of 1,267,558 image samples, including offline-augmented images, with 1,033,947 samples for training and 233,611 samples for testing. Fig.~\ref{dataset-statistics} presents detailed statistics of the dataset from multiple perspectives, including geographic distribution across regions, category-wise sample distribution, and pixel-level resolution distribution. These statistics offer an intuitive overview of the dataset structure and input characteristics, and serve as a foundation for subsequent model design and experimental analysis.

\subsubsection{\textbf{Challenge-Oriented Evaluation Sets}}
\begin{figure}[t]
    \centering
    \includegraphics[width=0.5\textwidth]{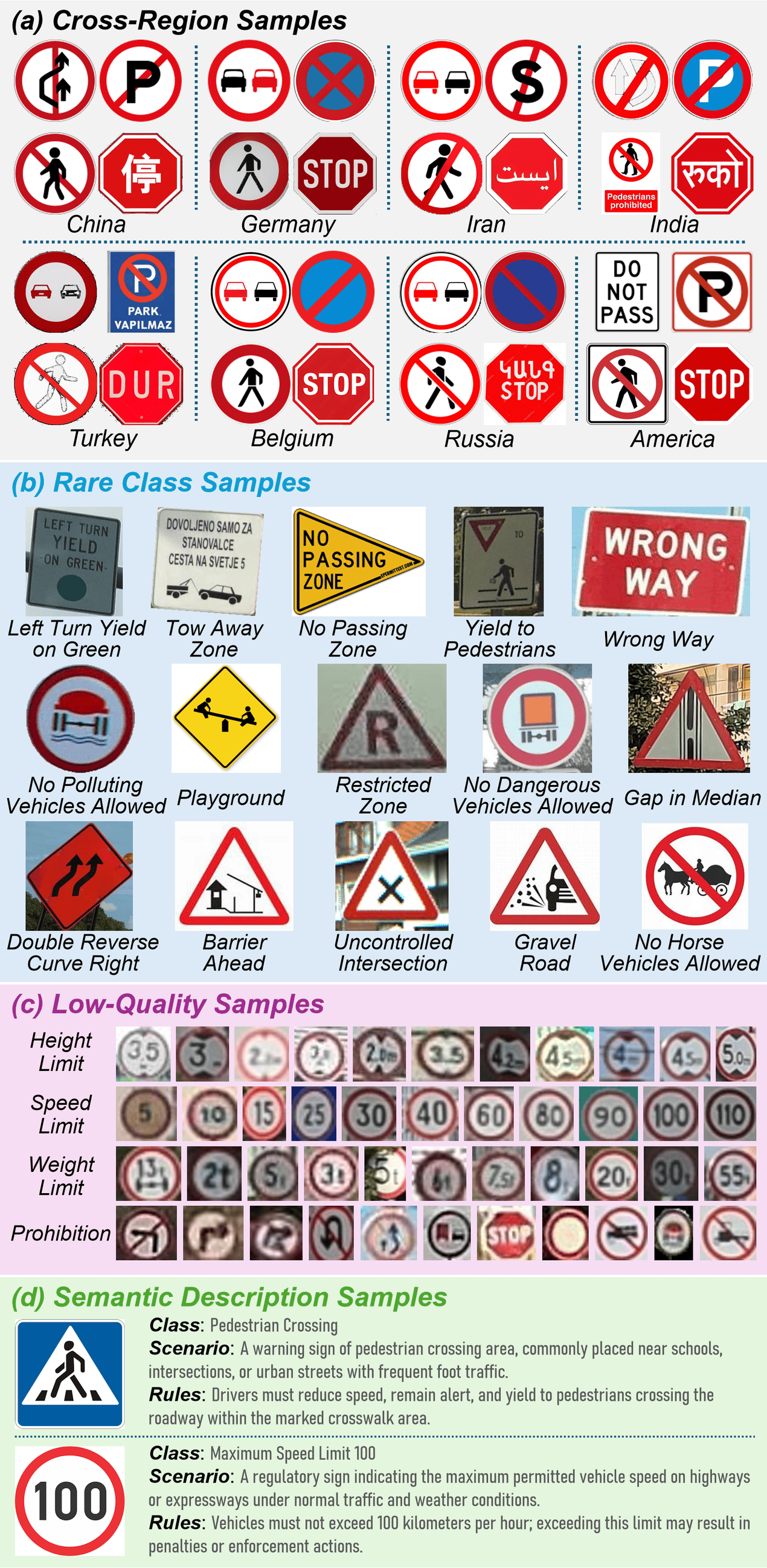}
    \vspace{-12pt}
    \caption{\textbf{Examples of challenge-oriented sets in TS-1M.} 
The figure illustrates representative samples from the cross-region, rare-class, low-clarity, and semantic description sets designed for systematic TSR evaluation.}
    \label{dataset-sample}
    \vspace{-15pt}
\end{figure}

To facilitate systematic evaluation of key challenges in modern TSR, we construct task-oriented evaluation sets with training subsets defined for each evaluation regime. In-domain challenge sets are drawn from the TS-1M test split. Transfer evaluation uses held-out target datasets and target-specific training subsets, with evaluation images excluded from the corresponding training data. Representative examples from each set are shown in Fig.~\ref{dataset-sample}. The evaluation sets are summarized as follows:

\textbf{Cross-Region and Cross-Dataset Recognition Sets}:  
Transfer is evaluated on ten public traffic sign datasets using target-specific holdouts. For each geographically localized target, we exclude the entire target dataset and all training images from its corresponding region across the remaining TS-1M sources before fine-tuning. The model is then evaluated on the held-out target test data. For the globally distributed MTSD, we exclude the entire MTSD source dataset from fine-tuning and evaluate cross-dataset transfer using its test split. The training subset for each transfer experiment is determined by these target-specific exclusions. Category definitions are aligned to the TS-1M taxonomy.

\textbf{Rare-Class Evaluation Set}:  
The two subsets contain test images from categories with fewer than 50 test samples (\emph{Rare $<$ 50}) and fewer than 40 test samples (\emph{Rare $<$ 40}). Because all categories use the same train/test split ratio, low test frequencies correspond to limited training samples for the same categories. These subsets therefore assess recognition of classes that are underrepresented during training. All evaluation images come from the TS-1M test split.

\textbf{Low-Clarity Evaluation Set}:  
This set contains traffic sign crops with resolutions below $30 \times 30$ pixels before model-input resizing. This fixed threshold selects small crops with limited spatial detail, as commonly encountered when distant signs occupy few pixels in full-scene images. It assesses recognition when fine symbols, digits, and boundaries are poorly resolved. All samples are drawn from the TS-1M test split.

\textbf{Semantic Description Set}:  
Unlike the image-based evaluation sets, this set provides structured textual descriptions for each of the 454 traffic sign categories in TS-1M, without visual samples. For each category, three types of semantic descriptions are included: \emph{scenario}, \emph{rules}, and \emph{scenario + rules}, constructed based on official traffic regulations and standardized interpretation guidelines. This set supports experiments on how scene and rule descriptions affect VLM classification during training and inference.

Together, these challenge-oriented evaluation sets enable fine-grained analysis of model strengths and limitations across multiple dimensions of TSR. The final release of TS-1M includes the complete core dataset, standardized category annotations, and optional semantic descriptions. As an open-source and extensible foundational dataset, TS-1M aims to provide a unified evaluation basis for the TSR community, facilitate systematic analysis of model generalization, and accelerate research on multimodal, cross-domain, and task-oriented traffic sign understanding.

\begin{figure*}[t!]
    \centering
    \includegraphics[width=1.0\textwidth]{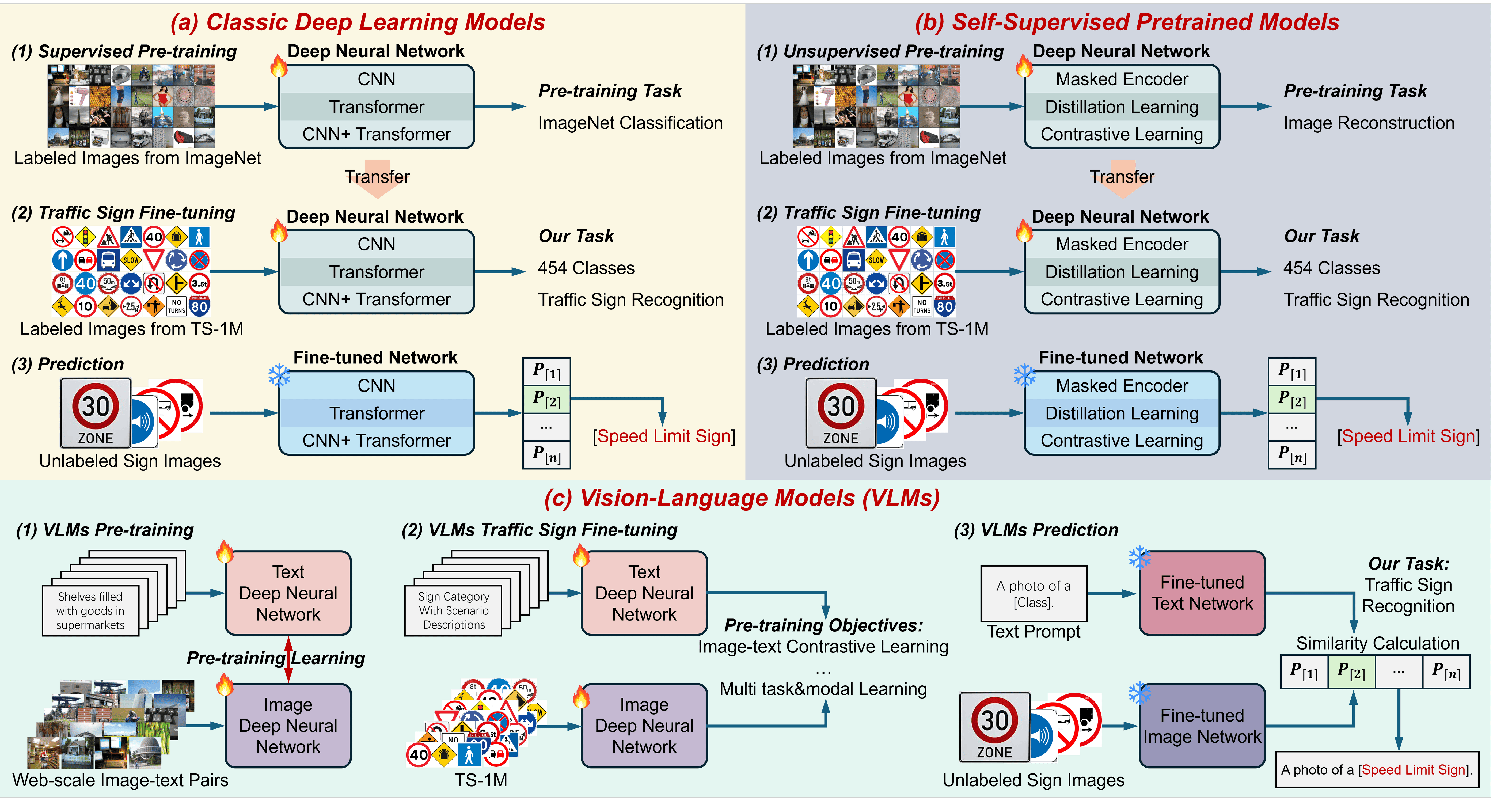}
    \vspace{-14pt}
    \caption{\textbf{Modeling frameworks of three representative learning paradigms in TS-1M.} 
The figure compares classic supervised models, self-supervised pretrained models, and VLMs from the perspectives of training pipelines and inference mechanisms.}
    \label{modeling-framework}
    \vspace{-10pt}
\end{figure*}

\section{Benchmark Modeling}
\label{sec:modeling}

\subsection{Overview of Modeling}

To systematically examine how different learning paradigms address the challenges of TSR, we establish a unified modeling benchmark grounded in the TS-1M dataset. The benchmark spans diverse model categories and design philosophies, enabling principled comparison across architectures under realistic TSR scenarios and supporting analysis of generalization behavior and deployment-oriented trade-offs.

We categorize the evaluated models into three representative groups, whose architectural differences are illustrated in Fig.~\ref{modeling-framework}. 
(a) \textbf{Classic Deep Learning Models}, which rely on supervised training and include convolutional and transformer-based visual backbones;  
(b) \textbf{Self-Supervised Pretrained Models}, which learn transferable visual representations from large-scale unlabeled data through pretext objectives and are adapted to downstream TSR tasks;
(c) \textbf{Vision-Language Models}, which jointly model visual inputs and language descriptions, enabling semantic alignment and enhanced generalization beyond purely visual cues.

These model classes reflect fundamentally different inductive biases and learning strategies. Classic CNNs emphasize local texture extraction and are typically more efficient, while transformer-based backbones enable long-range dependency modeling. Self-supervised models benefit from representation richness acquired through pretext tasks, improving generalization under data-sparse or long-tailed distributions. VLMs go beyond visual appearance by aligning image features with semantic cues, demonstrating potential in multilingual and text-centric sign interpretation.

To support systematic modeling analysis, Table~\ref{tab:model-types-comparison} provides a theoretical comparison of different model types across seven key dimensions. Four of these dimensions correspond to task-oriented challenges central to real-world TSR, including cross-region recognition, rare sign classification, low-clarity robustness, and recognition with semantic descriptions. The remaining three dimensions characterize architectural and practical properties of the models, namely receptive field modeling, data dependency, and deployment suitability. This comparison reflects structural characteristics and insights from prior literature, rather than empirical measurements. Quantitative validation of these modeling hypotheses is presented in Section~\ref{sec:experiments}.

\subsection{Modeling of Classic Models}
\label{sec:modeling:classic}

Classic deep learning models have long been foundational for TSR due to their architectural maturity, convergence stability, and deployment efficiency. In this benchmark, we evaluate a diverse set of representative classic architectures, including Standard CNNs (ResNet50/101~\cite{he2016deep}, ResNeXt50~\cite{xie2017aggregated}), Lightweight CNNs (ShuffleNetV2~\cite{ma2018shufflenet}, MobileNetV3~\cite{howard2019searching}, EfficientNetV2~\cite{tan2021efficientnetv2}), Hybrid CNN-Transformer (MobileViT~\cite{mehta2021mobilevit}, EdgeNeXt~\cite{maaz2022edgenext}), and Pure Transformer and Modernized CNNs (Vision Transformer~\cite{dosovitskiy2020image}, ConvNeXt~\cite{liu2022convnet}). 
Our goal is to analyze how their architectural designs influence performance across distinct TSR challenges. Table~\ref{tab:unified-model-comparison} presents the unified comparison across all model types, summarizing their architectural types, parameter sizes, and theoretical advantages within the TSR benchmark.

\subsubsection{\textbf{Training and Adaptation Strategy}}

To ensure fair comparison, all models are trained using a unified pipeline. Input images are resized to $224 \times 224$, initialized with ImageNet-pretrained weights, and fully fine-tuned on the TS-1M dataset. For Transformer-like architectures (e.g., MobileViT, ConvNeXt), we disable weight decay for LayerNorm parameters to stabilize training. Lightweight models (e.g., ShuffleNetV2, MobileNetV3) retain their original structure to preserve deployment efficiency.

\subsubsection{\textbf{Architectural Analysis and Theoretical Insights}}

From a mathematical perspective, CNNs extract hierarchical features via localized convolutional operations defined as:
\begin{equation}
f_{i,j}^{(l)} = \sigma \left( \sum_{m,n} W_{m,n}^{(l)} \cdot f_{i+m,j+n}^{(l-1)} + b^{(l)} \right),
\label{eq:cnn}
\end{equation}
where $f_{i,j}^{(l)}$ denotes the activation at location $(i,j)$ in layer $l$, $W^{(l)}$ is the convolutional kernel, $b^{(l)}$ is the bias, and $\sigma$ is the non-linear activation function. This inductive bias toward locality makes CNNs efficient for structured inputs like centered traffic signs but limits their global context modeling.

Transformers, by contrast, rely on self-attention mechanisms to capture global dependencies. The core operation is:
\begin{equation}
\mathrm{Attention}(Q, K, V) = \mathrm{softmax} \left( \frac{QK^T}{\sqrt{d_k}} \right)V,
\label{eq:attention}
\end{equation}
where $Q$, $K$, and $V$ are query, key, and value matrices derived from input embeddings, and $d_k$ is the dimensionality of the keys. This formulation enables long-range modeling of spatial patterns in traffic signs, which is beneficial for recognizing style-variant or context-rich signs.

Hybrid models such as MobileViT and EdgeNeXt attempt to balance both paradigms by injecting Transformer blocks into CNN backbones. This fusion enhances receptive field size without incurring significant computational cost.

\subsubsection{\textbf{Model Taxonomy and Capability Mapping}}
From a family-level perspective, classic models exhibit distinct strengths and limitations for TSR. 
\textbf{Standard CNNs} provide stable performance and are well-suited for structured, high-resolution signs due to their strong locality bias. \textbf{Lightweight CNNs} prioritize efficiency and are particularly attractive for real-time or edge deployment, although their reduced capacity can make them more sensitive to blur or ambiguity. \textbf{Hybrid CNN-Transformer} models balance local detail preservation with global context modeling, improving robustness to intra-class variation and cross-domain shifts. \textbf{Pure Transformer and Modernized CNNs} architectures benefit from global receptive fields and large-scale pretraining, showing advantages in rare sign recognition and diverse visual styles, at the cost of increased computational complexity.

\begin{table*}[t!]
\renewcommand\arraystretch{1.3}
\caption{\textbf{Theoretical Capability Analysis of Different Model Categories Across Critical TSR Dimensions.} This table reflects modeling hypotheses based on architecture design and prior literature. The empirical evaluation is provided in Section~\ref{sec:experiments}.}
\centering
\setlength{\tabcolsep}{1.8mm}
\footnotesize
\begin{tabular}{lccccccc}
\hline
\rowcolor{ts1m_purple!7} \textbf{Model Type} & \textbf{Cross-Region} & \textbf{Rare Signs} & \textbf{Low Clarity} & \textbf{Semantic Text} & \textbf{Receptive Field} & \textbf{Data Dependency} & \textbf{Deployment} \\
\hline
CNNs & \Medium Medium & \Medium Medium & \Medium Medium & \Low Low & \Medium Medium & \Medium Medium & \High High \\
\rowcolor{gray!9} Transformers& \High High & \High High & \Medium Medium & \Low Low & \Excellent Very High & \High High & \Medium Medium \\
Self-Supervised Models & \Excellent Very High & \Excellent Very High & \Excellent Very High & \Low Low & \Excellent Very High & \Low Low & \Medium Medium \\
\rowcolor{gray!9} Vision-Language Models & \Excellent Very High & \Excellent Very High & \High High & \Excellent Very High & \Excellent Very High & \Excellent Very High & \Low Low \\
\hline
\end{tabular}
\label{tab:model-types-comparison}
\end{table*}

\subsection{Modeling of Self-Supervised Pretrained Models}
\label{sec:modeling:selfsupervised}

SSL has emerged as an effective paradigm for visual representation learning, allowing models to acquire discriminative and transferable features from large-scale unlabeled data. This capability is particularly relevant to TSR, where domain shifts, rare categories, and visually degraded samples often undermine the performance of fully supervised approaches. 
In this benchmark, we evaluate five representative SSL methods, including MAE~\cite{he2022masked}, DINO~\cite{caron2021emerging}, MoCoV3~\cite{chen2021empirical}, SimCLR~\cite{chen2020simple}, and SimMIM~\cite{xie2022simmim}. MAE and SimMIM use masked image modeling, SimCLR and MoCoV3 use contrastive learning, and DINO uses self-distillation. All models are pretrained on ImageNet and subsequently fine-tuned on TS-1M for downstream TSR evaluation. A detailed comparison of their architectural characteristics is provided in Table~\ref{tab:unified-model-comparison}.

\subsubsection{\textbf{Training and Adaptation Strategy}}

All SSL models use an input resolution of $224 \times 224$ and are fine-tuned on TS-1M. SimCLR and MoCoV3 use ResNet-50 backbones with 454-class linear classification heads, with both the backbone and head jointly optimized. DINO uses a ViT-B/16 backbone with a linear classification head. MAE and SimMIM are adapted using their visual encoders and classification heads, with reconstruction decoders excluded from downstream classification.

\subsubsection{\textbf{Theoretical Foundations and Model Insights}}

\textbf{MAE} and \textbf{SimMIM} rely on masked image modeling. A subset of patches $\mathbf{x}_m \subset \mathbf{x}$ is masked, and the model learns to reconstruct them based on visible tokens $\mathbf{x}_{\setminus m}$. For MAE, the reconstruction objective is:
\begin{equation}
\mathcal{L}_{\text{MIM}} = \left\| f_{\text{dec}}(f_{\text{enc}}(\mathbf{x}_{\setminus m})) - \mathbf{x}_m \right\|_2^2.
\label{eq:mim}
\end{equation}
This encourages the encoder to model spatial structure and contextual relations.

\textbf{SimCLR} is a pure contrastive learning framework. For two augmented views $(\mathbf{z}_i, \mathbf{z}_j)$ of the same image, its InfoNCE loss is:
\begin{equation}
\mathcal{L}_{\text{SimCLR}} = -\log 
\frac{\exp(\text{sim}(\mathbf{z}_i, \mathbf{z}_j)/\tau)}
{\sum_{k} \exp(\text{sim}(\mathbf{z}_i, \mathbf{z}_k)/\tau)},
\label{eq:simclr}
\end{equation}
where $\tau$ is a temperature parameter. This objective strengthens instance-level discrimination across augmented views.

\textbf{MoCoV3} extends contrastive learning with a momentum-updated encoder to maintain consistency in negative samples. Given an anchor $\mathbf{q}$, positive key $\mathbf{k}_+$, and negatives $\{\mathbf{k}_-\}$:
\begin{equation}
\mathcal{L}_{\text{MoCo}} = -\log 
\frac{\exp(\mathbf{q}\cdot\mathbf{k}_+/\tau)}
{\sum_{\mathbf{k}\in\{\mathbf{k}_+,\mathbf{k}_-\}} \exp(\mathbf{q}\cdot\mathbf{k}/\tau)}.
\label{eq:moco}
\end{equation}
The momentum encoder provides targets for contrastive representation learning.

\textbf{DINO} learns visual representations through self-distillation without labels. Given teacher and student outputs $p^{(t)}$ and $p^{(s)}$, the loss is:
\begin{equation}
\mathcal{L}_{\text{DINO}}
= -\sum_{i} p_i^{(t)} \log p_i^{(s)}.
\label{eq:dino}
\end{equation}
This encourages agreement between teacher and student predictions across augmented views.

\subsubsection{\textbf{Model Suitability and Summary}}

These SSL methods provide complementary representation learning strategies:

\textbf{MAE/SimMIM}: Masked image modeling learns spatial structure from visible context.
\textbf{SimCLR}: Contrastive learning promotes instance-level discrimination across augmented views.
\textbf{MoCoV3}: A momentum encoder provides targets for contrastive learning.
\textbf{DINO}: Self-distillation aligns teacher and student predictions across augmented views.

\begin{table*}[t!]
\renewcommand\arraystretch{1.3}
\caption{\textbf{Unified Comparison of Benchmarked Models for TSR.} The descriptions emphasize structural and learning characteristics rather than task-specific performance categories.} 
\centering
\setlength{\tabcolsep}{1.5mm}
% \footnotesize
\scriptsize
\begin{tabular}{llll}
\hline
\rowcolor{ts1m_purple!7} \textbf{Model} & \textbf{Model Variant} & \textbf{Pretraining Type / Structure} & \textbf{Key Characteristics for TSR Task} \\
\hline
ResNet50 \cite{he2016deep}    &   ResNet50  & Standard CNN      & Robust baseline with fast inference and stable performance on large, clean signs \\
\rowcolor{gray!9} ResNet101 \cite{he2016deep}    &  ResNet101  & Standard CNN           & Higher-capacity backbone; improved precision under well-lit, structured scenes \\
ResNeXt50 \cite{xie2017aggregated}    &  ResNext50  & Standard CNN         & Flexible feature encoding for fine-grained visual distinctions \\
\rowcolor{gray!9} ShuffleNetV2 \cite{ma2018shufflenet}  &  ShuffleNetV2  & Lightweight CNN             & Extremely efficient, ideal for embedded or low-power roadside devices \\
MobileNetV3 \cite{howard2019searching}   & MobileNetV3 Small & Lightweight CNN      & Well-balanced speed-accuracy design; robust for on-vehicle deployment \\
\rowcolor{gray!9} EfficientNetV2 \cite{tan2021efficientnetv2} & EfficientNetV2 Small  & Scalable CNN            & High efficiency under varying sign sizes via compound scaling \\
MobileViT \cite{mehta2021mobilevit}  & MobileViT Small & Hybrid CNN+Transformer    & Effective local-global fusion for dense urban scenes and varied backgrounds \\
\rowcolor{gray!9} EdgeNeXt \cite{maaz2022edgenext}  &  EdgeNext Base  & Hybrid CNN+Transformer    & Strong edge-aware representation; performs well on low-quality or blurred signs \\
Vision Transformer \cite{dosovitskiy2020image}  & ViT-Base/16 & Pure Transformer    & Strong global context modeling; useful for complex intra-class variability \\
\rowcolor{gray!9} ConvNeXt \cite{liu2022convnet}  & ConvNext Small & Modernized CNN   & Transformer-inspired CNN; stable across style-diverse visual domains \\
\hline
MAE \cite{he2022masked}  & ViT-Base/16 & Masked Image Modeling & Masked-patch reconstruction with a ViT encoder for visual pretraining \\
\rowcolor{gray!9} SimMIM \cite{xie2022simmim} & Swin-Base/16 & Masked Image Modeling & Masked image prediction with a hierarchical Swin Transformer backbone \\
DINO \cite{caron2021emerging} & ViT-Base/16 & Self-Distillation & Teacher-student prediction consistency across multiple augmented image views \\
\rowcolor{gray!9} MoCoV3 \cite{chen2021empirical} & ResNet50 & Contrastive Learning & Contrastive visual pretraining with a momentum-updated target encoder \\
SimCLR \cite{chen2020simple} & ResNet50 & Contrastive Learning & Instance-level visual discrimination across multiple augmented image views \\
\hline
\rowcolor{gray!9} CLIP \cite{radford2021learning} & ViT-B/16 & Contrastive Learning & Strong open-vocabulary alignment; effective for text-bearing or symbolic signs \\
BLIP \cite{li2022blip}   & ViT-Base/16 & Multi-task Vision-Language Fusion   & ITC candidate retrieval followed by ITM reranking for traffic sign classification \\
\rowcolor{gray!9} BLIP2 \cite{li2023blip}  & ViT-g/14 + OPT-2.7B & Frozen ViT + Q-Former + LLM & Generates traffic sign class names from visual features \\
LLaVA \cite{liu2023visual}  & LLaVA-v1.5-7b & Projector-tuned VLM & Class-name generation with frozen vision and language backbones \\
\hline
\end{tabular}
\vspace{-10pt}
\label{tab:unified-model-comparison}
\end{table*}

\subsection{Modeling of Vision-Language Models}
\label{sec:modeling:vlm}

VLMs have recently reshaped multimodal perception by learning aligned semantic representations between images and natural language. Their ability to leverage textual priors, interpret symbolic content, and generalize to open-vocabulary categories is particularly beneficial for TSR, where many signs contain numbers, textual elements, or rule-dependent semantics. 
In this benchmark, we evaluate four representative VLMs, including CLIP~\cite{radford2021learning}, BLIP~\cite{li2022blip}, BLIP2~\cite{li2023blip}, and LLaVA~\cite{liu2023visual}. These models span contrastive vision-language alignment, multimodal fusion, and vision-conditioned large language modeling paradigms. All models are adapted to TS-1M under common data splits and evaluation metrics, using model-specific adaptation and inference procedures. Architectural characteristics and comparative properties of these models are summarized in Table~\ref{tab:unified-model-comparison}.

\subsubsection{\textbf{Training and Adaptation Strategy}}

All VLMs are initialized from publicly available pretrained checkpoints and adapted using model-specific trainable modules. They share the TS-1M category definitions and evaluation metrics; category texts and instruction prompts follow each model's prediction formulation. Implementation details are provided in the Supplementary Material.

CLIP uses a ViT-B/16 visual backbone and predicts the category with the highest image-text cosine similarity. BLIP uses image-text contrastive (ITC) retrieval followed by image-text matching (ITM) reranking. BLIP-2 retains a frozen visual encoder and generates category text, with predictions obtained by greedy decoding and evaluated by string matching. LLaVA keeps both the vision encoder and language model frozen, trains only the multimodal projector, and generates the traffic sign class name for evaluation.

\subsubsection{\textbf{Theoretical Foundations and Model Insights}}

The four VLMs differ in their architectural designs and learning objectives, providing distinct mechanisms for image-text matching and category-text generation.

\textbf{CLIP} adopts a dual-encoder architecture that independently encodes images and text into a shared embedding space. Given an image $x$ and its corresponding textual prompt $t$, CLIP aligns their embeddings $f(x)$ and $g(t)$ via a contrastive learning objective:
\begin{equation}
\mathcal{L}_{\text{CLIP}} = -\log 
\frac{\exp(\text{sim}(f(x), g(t))/\tau)}
{\sum_{t'} \exp(\text{sim}(f(x), g(t'))/\tau)},
\label{eq:cliploss}
\end{equation}
where $\text{sim}(\cdot,\cdot)$ denotes cosine similarity and $\tau$ is a learnable temperature parameter. This formulation encourages semantic alignment at the instance level and enables open-vocabulary recognition. At inference, image and category-text embeddings are L2-normalized and compared by cosine similarity to obtain the predicted class.

\textbf{BLIP} combines ITC retrieval with cross-attention-based ITM reranking for TSR classification. Given a test image, the ITC branch computes its similarity to all 454 category texts and retrieves the top $K=50$ candidates. The ITM head then reranks these candidates, and the top-ranked category is returned as the prediction. Candidate selection and reranking use only the input image and predefined category texts; ground-truth test labels are used solely to compute evaluation metrics. The ITM training objective is defined as:
\begin{equation}
\mathcal{L}_{\text{ITM}} = 
- y \log P(\text{match}|x,t)
- (1-y)\log P(\text{non-match}|x,t),
\label{eq:blipitm}
\end{equation}
where $y \in \{0,1\}$ is the match label of a training image-text pair. This objective trains the ITM head to assess fine-grained cross-modal consistency beyond the global similarity used for candidate retrieval.

\textbf{BLIP-2} decouples visual representation learning and language modeling by introducing a lightweight Querying Transformer (Q-Former) that bridges a frozen vision encoder and a large pretrained language model. Given visual features $\mathbf{v}$ extracted from the image encoder, the Q-Former produces a compact set of language-aligned queries:
\begin{equation}
\mathbf{q} = f_{\text{QF}}(\mathbf{v}),
\label{eq:qformer}
\end{equation}
which are then projected into the LLM embedding space to condition category-text generation. In our TSR adaptation, the visual encoder remains frozen. The language model generates the category name through greedy decoding, and the decoded text is matched against class names to compute classification metrics.

\textbf{LLaVA} further extends vision-language modeling into an instruction-following paradigm by integrating a visual encoder with an instruction-tuned LLM. Visual features $\mathbf{h}_v$ and textual token embeddings $\mathbf{h}_t$ are concatenated and processed by a multimodal transformer:
\begin{equation}
\mathbf{H} = \text{Transformer}([\mathbf{h}_v ; \mathbf{h}_t]),
\label{eq:llava}
\end{equation}
allowing the model to perform high-level multimodal reasoning. This formulation implicitly captures rule-based and contextual semantics, which are common in regulatory and warning traffic signs. For TSR adaptation, only the multimodal projector is trained while the vision encoder and language model remain frozen. At inference, the model autoregressively generates a traffic sign class name, which is matched against class names for evaluation.

\subsubsection{\textbf{Model Suitability and Summary}}

Each VLM paradigm exhibits complementary strengths for TSR:

\textbf{CLIP}: Strong open-vocabulary alignment and robustness to unseen textual descriptions, suitable for scalable and extensible TSR settings.  
\textbf{BLIP}: ITC retrieval and ITM reranking combine efficient candidate selection with fine-grained image-text matching.  
\textbf{BLIP-2}: Efficient semantic abstraction via Q-Former enables high-level reasoning with reduced computational overhead.  
\textbf{LLaVA}: Instruction-driven multimodal reasoning provides potential advantages for context-dependent and rule-oriented traffic sign interpretation.

\section{Experiment and Results}
\label{sec:experiments}

\subsection{Experimental Settings}
All models were implemented in PyTorch and trained on four NVIDIA A100-PCIE-40GB GPUs, with inference conducted on a single RTX 5090 GPU to ensure consistent evaluation. Unless otherwise specified, input images were resized to $224 \times 224$ and normalized following ImageNet conventions. Classical and self-supervised baselines were initialized from their respective official supervised or self-supervised checkpoints and fine-tuned utilizing the \texttt{mmpretrain} framework. To ensure optimal convergence, their training epochs (30–50), optimizers (SGD or AdamW), and learning rate schedules were strictly tailored to each specific architecture. VLMs were adapted for 20 epochs using model-specific objectives and trainable modules. CLIP and BLIP predict from image-text scores, while BLIP-2 and LLaVA generate category text. Comprehensive hyperparameter configurations are detailed in the Supplementary Material.

All downstream training images are drawn from TS-1M. We evaluate models under two regimes. (i) In-domain evaluation uses the core TS-1M training split and evaluates on its test split, including the rare-class, low-clarity, and semantic evaluation settings. (ii) Cross-region and cross-dataset evaluation uses a separate training run for each target dataset. Its fine-tuning pool excludes the target source dataset and, for geographically localized targets, all images from the same region in other TS-1M sources. All models use the same retained training data for a given target. Consequently, each column of Table~\ref{TableCross} corresponds to a target-specific training run. The holdout protocol is detailed in the Supplementary Material.

We report Top-1 accuracy as the primary metric, complemented by macro-precision, macro-recall, and macro F1-score to account for class imbalance. In Table~\ref{TableCross}, Avg.\ is weighted by the number of evaluated test images in each target dataset. The $\Delta$ column reports the absolute Top-1 accuracy difference from ResNet50 under the same evaluation setting, expressed in percentage points.

BLIP is evaluated as a complete ITC retrieval and ITM reranking pipeline. Retrieval considers the full category vocabulary, and all test images contribute to the reported classification metrics, including those whose correct class is not retrieved among the top $K=50$ candidates.

\begin{table*}[t!]
\renewcommand\arraystretch{1.3}
% \caption{Overall experiment on TS-1M dataset.}
\caption{\textbf{Overall Benchmark Results on the TS-1M Dataset.} 
The table compares model performance and computational cost across classic supervised models, self-supervised pretrained models, and VLMs. 
$\Delta$ indicates the Acc.@1 relative improvement with respect to the ResNet50 baseline. \best{} and \second{} highlight the best and second-best results, respectively.}
\vspace{-3pt}
\centering
\setlength{\tabcolsep}{3.0mm}
% \footnotesize
\scriptsize
% \begin{tabular}{>{\centering\arraybackslash}m{2.2cm}c>{\centering\arraybackslash}m{1.8cm}cccccc}
\begin{tabular}{llccccccc}
\hline 
\textbf{Method} & \textbf{Model} & \textbf{Params (M)$\downarrow$} & \textbf{FLOPs (G)$\downarrow$} & \textbf{Acc.@1$\uparrow$} & \textbf{Pre.$\uparrow$} & \textbf{Recall$\uparrow$} & \textbf{F-1$\uparrow$} & \textbf{$\Delta$ (\%)$\uparrow$} \\
\hline
\multirow{10}{*}{Classic Model} 
& ResNet50 \cite{he2016deep}       & 25.6  & 4.09  & 0.9260 & 0.8724 & 0.8633 & 0.8570 & - \\
& ResNet101 \cite{he2016deep}      & 44.5  & 7.80  & 0.9315 & 0.8899 & 0.8979 & 0.8844 & +0.55 \\
& ResNext50 \cite{xie2017aggregated} & 25.0  & 4.23  & 0.9403 & 0.9005 & 0.9007 & 0.8920 & +1.43 \\
& ShufflenetV2 \cite{ma2018shufflenet} & \cellcolor[HTML]{FFF2CC}\textbf{2.28}  & \cellcolor[HTML]{F1E9DF}0.15  & 0.9234 & 0.8583 & 0.8811 & 0.8592 & -0.26 \\
& MobileNetV3 \cite{howard2019searching} & \cellcolor[HTML]{F1E9DF}2.54  & \cellcolor[HTML]{FFF2CC}\textbf{0.06}  & 0.9265 & 0.8876 & 0.8653 & 0.8628 & +0.05 \\
& EfficientNetV2 \cite{tan2021efficientnetv2} & 21.46 & 9.72  & 0.9372 & 0.9135 & 0.8954 & 0.8977 & +1.12 \\
& MobileViT \cite{mehta2021mobilevit} & 5.58  & 2.03  & 0.9277 & 0.8855 & 0.8888 & 0.8786 & +0.17 \\
& EdgeNeXt \cite{maaz2022edgenext} & 18.51 & 3.81  & 0.9369 & 0.9033 & 0.9021 & 0.8951 & +1.09 \\
& Vision Transformer \cite{dosovitskiy2020image} & 86.56 & 17.58 & 0.9244 & 0.9188 & 0.8478 & 0.8658 & -0.16 \\
& ConvNeXt \cite{liu2022convnet} & 50.22 & 8.69  & 0.9428 & 0.9146 & \cellcolor[HTML]{FFF2CC}\textbf{0.9086} & \cellcolor[HTML]{F1E9DF}0.9043 & +1.68 \\
\hline
\multirow{5}{*}{Pretrained Model}
& MAE \cite{he2022masked}    & 86.56 & 17.58 & 0.9160 & 0.8733 & 0.8700 & 0.8575 & -1.00 \\
& SimMIM \cite{xie2022simmim} & 87.77 & 15.14  & 0.9403 & \cellcolor[HTML]{F1E9DF}0.9236 & 0.8935 & 0.8972 & +1.43 \\
& DINO \cite{caron2021emerging} & 86.56 & 17.58 & 0.9190 & 0.8898 & 0.8516 & 0.8595 & -0.70 \\
& MoCoV3 \cite{chen2021empirical} & 25.6  & 4.09 & 0.9385 & 0.9008 & 0.8935 & 0.8876 & +1.25 \\
& SimCLR \cite{chen2020simple} & 25.6  & 4.09  & 0.9364 & 0.8989 & 0.8929 & 0.8868 & +1.04 \\
\hline
\multirow{4}{*}{VLM} 
& CLIP \cite{radford2021learning} & 149.6 & 39.7 & \cellcolor[HTML]{FFF2CC}\textbf{0.9533} & \cellcolor[HTML]{FFF2CC}\textbf{0.9310} & \cellcolor[HTML]{F1E9DF}0.9084 & \cellcolor[HTML]{FFF2CC}\textbf{0.9113} & \cellcolor[HTML]{FFF2CC}\textbf{+2.73} \\
& BLIP \cite{li2022blip} & 223.7 & 67.2 & \cellcolor[HTML]{F1E9DF}0.9448 & 0.8872 & 0.8949 & 0.8770 & \cellcolor[HTML]{F1E9DF}+1.88 \\
& BLIP2 \cite{li2023blip} & 3756.2 &  611.8 & 0.9432 & 0.8828 & 0.8883 & 0.8776 & +1.72 \\
& LLAVA \cite{liu2023visual} & 7062.9 & 6216.7 & 0.9406 & 0.9136 & 0.8677 & 0.8776 & +1.46 \\
\hline
\end{tabular}
\label{TableAll}
% \vspace{-2pt}
\end{table*}

\subsection{Overall Benchmark Results on TS-1M}

Table~\ref{TableAll} reports the overall performance of all benchmarked models on the TS-1M test set. Among classic architectures, modern convolutional networks achieve the highest accuracy, with ConvNeXt reaching 0.9428 (+1.68\%) and models such as ResNeXt50, EfficientNetV2, and EdgeNeXt also showing consistent improvements over the ResNet50 baseline. In contrast, lightweight CNNs and the standard ViT model perform less favorably, suggesting that the strong inductive bias of convolutional structures remains advantageous for the structured visual patterns of traffic signs.

Among self-supervised pretrained models, SimMIM achieves the highest accuracy (0.9403) and macro-F1 score (0.8972). MoCoV3 and SimCLR also exceed the ResNet50 baseline in accuracy by 1.25 and 1.04 percentage points, respectively, whereas MAE and DINO obtain lower accuracy than the baseline.

VLMs achieve the best overall performance. CLIP attains the highest accuracy of 0.9533 (+2.73\%) and leads across Precision, Recall, and F1-score, demonstrating the strong advantage of image-text alignment for categories containing numbers, symbols, and explicit semantic cues. BLIP, BLIP2, and LLAVA likewise surpass purely visual models, confirming that multimodal semantic priors substantially enhance large-scale TSR. Overall, these results show that modern CNNs remain competitive strong baselines, SSL models offer selective improvements depending on architecture, and VLMs set the new performance ceiling on TS-1M.

\begin{table*}[t]
\renewcommand\arraystretch{1.3}
\caption{\textbf{Cross-Region and Cross-Dataset Generalization under Target Holdout.} 
Top-1 accuracy is reported for each target dataset. Avg.\ is weighted by test image counts; $\Delta$ denotes the difference from ResNet50 in percentage points. \best{} and \second{} indicate the best and second-best results, respectively.}
\vspace{-3pt}
\centering
\setlength{\tabcolsep}{1.5mm}
\scriptsize
\begin{tabular}{llcccccccccccc}
\hline 
\multirow{2}{*}{\textbf{Method}} & \multirow{2}{*}{\textbf{Model}} 
& \textbf{TT100K}
& \textbf{GTSRB}
& \textbf{PTSD}
& \textbf{TurkeyTS}
& \textbf{PolishTS}
& \textbf{BelgiumTS}
& \textbf{RTSD}
& \textbf{MTSD}
& \textbf{DFG}
& \textbf{ARTS}
& \multirow{2}{*}{\textbf{Avg.}} & \multirow{2}{*}{\textbf{$\Delta$ (pp)}} \\
\cline{3-12}
&  & \textbf{China} & \textbf{Germany} & \textbf{Iran} & \textbf{Turkey} & \textbf{Poland} & \textbf{Belgium} & \textbf{Russia} & \textbf{Multi-region} & \textbf{Slovenia} & \textbf{USA} \\
\hline 
\multirow{10}{*}{\begin{tabular}[l]{@{}l@{}}Classic \\ Model\end{tabular}} 
& ResNet50 \cite{he2016deep} & 0.9255 & 0.9488 & 0.9143 & 0.9020 & 0.8461 & 0.9507 & 0.9306 & 0.8503 & 0.8495 & 0.9072 & 0.8900 & - \\
& ResNet101 \cite{he2016deep} & 0.9393 & \cellcolor[HTML]{F1E9DF}0.9492 & 0.9126 & 0.9050 & 0.8561 & 0.9502 & 0.9311 & 0.8497 & 0.8746 & 0.9325 & 0.8978 & +0.78 \\
& ResNext50 \cite{xie2017aggregated} & 0.9370 & 0.9447 & 0.9263 & 0.9340 & 0.8563 & 0.9551 & 0.9350 & 0.8636 & 0.9107 & 0.9508 & 0.9091 & +1.91 \\
& ShufflenetV2 \cite{ma2018shufflenet} & 0.9374 & 0.9474 & 0.9136 & 0.9118 & 0.8654 & 0.9454 & 0.9374 & 0.8272 & 0.8340 & 0.9263 & 0.8861 & -0.39 \\
& MobileNetV3 \cite{howard2019searching} & 0.9290 & 0.9448 & 0.9172 & 0.9141 & 0.8526 & 0.9400 & 0.9322 & 0.8470 & 0.8393 & 0.9148 & 0.8895 & -0.05 \\
& EfficientNetV2 \cite{tan2021efficientnetv2} & 0.9409 & 0.9417 & 0.9122 & 0.9184 & 0.8592 & 0.9556 & 0.9414 & 0.8553 & 0.8949 & 0.9434 & 0.9036 & +1.35 \\
& MobileVIT \cite{mehta2021mobilevit} & 0.9378 & 0.9455 & 0.9050 & 0.9134 & 0.8552 & 0.9417 & 0.9355 & 0.8412 & 0.8598 & 0.9316 & 0.8927 & +0.27 \\
& EdgeNeXt \cite{maaz2022edgenext} & 0.9337 & 0.9417 & 0.9184 & 0.9176 & 0.8470 & 0.9493 & 0.9416 & 0.8610 & 0.8926 & 0.9374 & 0.9031 & +1.31 \\
& Vision Transformer \cite{dosovitskiy2020image} & 0.9163 & 0.9477 & 0.9151 & 0.9106 & 0.8551 & 0.9489 & 0.9326 & 0.8404 & 0.8855 & 0.8914 & 0.8860 & -0.40 \\
& ConvNeXt \cite{liu2022convnet} & \cellcolor[HTML]{F1E9DF}0.9442 & 0.9441 & 0.9185 & 0.9301 & 0.8618 & 0.9530 & \cellcolor[HTML]{F1E9DF}0.9425 & 0.8707 & \cellcolor[HTML]{F1E9DF}0.9302 & \cellcolor[HTML]{F1E9DF}0.9521 & \cellcolor[HTML]{F1E9DF}0.9142 & \cellcolor[HTML]{F1E9DF}+2.41 \\
\hline
\multirow{5}{*}{\begin{tabular}[l]{@{}l@{}}Pretrained \\ Model\end{tabular}}
& MAE \cite{he2022masked} & 0.9323 & 0.9454 & 0.9212 & 0.8980 & 0.8517 & 0.9572 & 0.9371 & 0.8118 & 0.8822 & 0.8816 & 0.8760 & -1.40 \\
& SimMIM \cite{xie2022simmim} & 0.9358 & 0.9426 & 0.9197 & 0.9114 & 0.8808 & 0.9503 & 0.9414 & 0.8626 & 0.9208 & 0.9414 & 0.9083 & +1.83 \\
& DINO \cite{caron2021emerging} & 0.9270 & 0.9477 & 0.9261 & 0.9129 & 0.8645 & 0.9516 & 0.9385 & 0.8097 & 0.8948 & 0.9260 & 0.8843 & -0.58 \\
& MoCoV3 \cite{chen2021empirical} & 0.9377 & 0.9413 & 0.9182 & 0.9326 & 0.8546 & 0.9488 & 0.9413 & 0.8661 & 0.8871 & 0.9397 & 0.9057 & +1.57 \\
& SimCLR \cite{chen2020simple} & 0.9390 & \cellcolor[HTML]{FFF2CC}\textbf{0.9497} & 0.9197 & 0.9304 & 0.8700 & \cellcolor[HTML]{F1E9DF}0.9593 & 0.9333 & 0.8576 & 0.8838 & 0.9369 & 0.9037 & +1.37 \\
\hline
\multirow{4}{*}{VLM} 
& CLIP \cite{radford2021learning} & \cellcolor[HTML]{FFF2CC}\textbf{0.9496} & 0.9432 & \cellcolor[HTML]{FFF2CC}\textbf{0.9448} & \cellcolor[HTML]{FFF2CC}\textbf{0.9770} & \cellcolor[HTML]{FFF2CC}\textbf{0.9719} & \cellcolor[HTML]{FFF2CC}\textbf{0.9615} & \cellcolor[HTML]{FFF2CC}\textbf{0.9447} & \cellcolor[HTML]{FFF2CC}\textbf{0.9058} & \cellcolor[HTML]{FFF2CC}\textbf{0.9444} & \cellcolor[HTML]{FFF2CC}\textbf{0.9545} & \cellcolor[HTML]{FFF2CC}\textbf{0.9351} & \cellcolor[HTML]{FFF2CC}\textbf{+4.51} \\
& BLIP \cite{li2022blip} & 0.9296 & 0.9395 & 0.9119 & 0.9032 & 0.8618 & 0.9342 & 0.9298 & 0.8698 & 0.8801 & 0.9145 & 0.8997 & +0.97 \\
& BLIP2 \cite{li2023blip} & 0.9210 & 0.9101 & 0.9352 & 0.9460 & 0.8857 & 0.9563 & 0.9385 & 0.8622 & 0.9138 & 0.9184 & 0.8999 & +0.98 \\
& LLAVA \cite{liu2023visual} & 0.9071 & 0.9009 & \cellcolor[HTML]{F1E9DF}0.9411 & \cellcolor[HTML]{F1E9DF}0.9727 & \cellcolor[HTML]{F1E9DF}0.9572 & 0.9518 & 0.9182 & \cellcolor[HTML]{F1E9DF}0.8743 & 0.9187 & 0.9172 & 0.9041 & +1.40 \\
\hline
\end{tabular}
\label{TableCross}
\vspace{-10pt}
\end{table*}

% \begin{tabular}[c]{@{}c@{}}Used \\ Classes\end{tabular}

\subsection{Cross-Region and Cross-Dataset Generalization}
\label{sec:transfer}
Table~\ref{TableCross} reports ten target-specific evaluations under the holdout protocol. For each regional target, training excludes both the target source dataset and its geographic region across the other sources. For the globally distributed MTSD target, the entire source dataset is excluded from fine-tuning for cross-dataset evaluation. Each model is fine-tuned on the corresponding retained training subset before evaluation on the target test split. Training membership is determined separately for each target by source provenance and the exclusion rules above.

TSCLIP~\cite{zhao2024tsclip} improves cross-region transfer through scene- and rule-aware prompts and adaptive dynamic weight ensembling. Complementing this method-level approach, TS-1M provides a larger multi-source data foundation and a broader category inventory for training and evaluating transfer across regions.

Among classic models, ConvNeXt achieves the highest accuracy on five of the ten targets, including TT100K (0.9442), MTSD (0.8707), DFG (0.9302), and ARTS (0.9521). Self-supervised models show target-dependent performance: SimMIM achieves 0.8808 on PolishTS and 0.9208 on DFG, while SimCLR attains the highest GTSRB accuracy of 0.9497. These results indicate that transfer performance varies with both the model and the target domain.

All evaluated models achieve test-image-weighted average accuracies above 87\% across the ten target datasets. These results suggest that TS-1M's large-scale, globally collected data and 454-category inventory provide a useful foundation for generalization across held-out regions and datasets. CLIP leads on nine of the ten targets, further highlighting the effectiveness of VLMs on this broader data foundation.

\subsection{Rare-Class Recognition} 

To assess recognition of categories with limited training representation, we evaluate all models on classes with fewer than 40 and 50 test samples under the common per-category split ratio (Table~\ref{TableRare}). Among classic models, ConvNeXt, EfficientNetV2, and EdgeNeXt achieve the strongest results, with ConvNeXt reaching the highest accuracies of 0.8784 and 0.8829 (+4.03\% and +3.91\%). These results indicate that modern convolutional architectures better capture fine-grained structures and subtle category differences under limited training representation, whereas lightweight models and the standard ViT struggle with reduced recall and F1-score, reflecting their limited robustness in long-tailed settings.

SimMIM achieves the highest SSL accuracy on both rare-class sets, exceeding ResNet50 by 3.25 and 2.92 percentage points, respectively. MoCoV3 and SimCLR also improve upon the baseline, whereas MAE and DINO obtain lower accuracy. VLMs exhibit the strongest overall performance, with CLIP achieving substantial gains of +9.10\% and +8.18\% in accuracy. This demonstrates that text-aligned semantic priors greatly enhance generalization to rare traffic sign categories.

\begin{table*}[t!]
\renewcommand\arraystretch{1.3}
\caption{\textbf{Performance on Rare-Class Recognition in the TS-1M Benchmark.} 
Results are reported on categories with fewer than 40 test samples (\textbf{Rare $<$ 40}) and fewer than 50 test samples (\textbf{Rare $<$ 50}) per class.
$\Delta$ denotes the Acc.@1 relative improvement over the ResNet50 baseline. \best{} and \second{} highlight the best and second-best results, respectively.}
\centering
% \footnotesize
\scriptsize
% \begin{tabular}{c>{\centering\arraybackslash}m{2.8cm}|ccccc|ccccc}
\begin{tabular}{ll|ccccc|ccccc}
\hline
\multirow{2}{*}{\textbf{Method}} & \multirow{2}{*}{\textbf{Model}}
 & \multicolumn{5}{c|}{\textbf{Rare $<$ 40}} 
 & \multicolumn{5}{c}{\textbf{Rare $<$ 50}} \\ 
\cline{3-12}
 &  & \textbf{Acc.@1} & \textbf{Pre.} & \textbf{Recall} & \textbf{F-1} & \textbf{$\Delta$ (\%)} & \textbf{Acc.@1} & \textbf{Pre.} & \textbf{Recall} & \textbf{F-1} & \textbf{$\Delta$ (\%)} \\
\hline \multirow{10}{*}{Classic Model} 
& ResNet50 \cite{he2016deep} & 0.8381 & 0.9460 & 0.8373 & 0.8636 & - & 0.8438 & 0.9469 & 0.8424 & 0.8683 & - \\
& ResNet101 \cite{he2016deep} & 0.8610 & 0.9532 & 0.8597 & 0.8767 & +2.29 & 0.8639 & 0.9521 & 0.8625 & 0.8792 & +2.01 \\
& ResNext50 \cite{xie2017aggregated} & 0.8530 & 0.9518 & 0.8525 & 0.8745 & +1.49 & 0.8601 & 0.9519 & 0.8568 & 0.8794 & +1.63 \\
& ShufflenetV2 \cite{ma2018shufflenet} & 0.8334 & 0.9303 & 0.8307 & 0.8509 & -0.47 & 0.8395 & 0.9281 & 0.8363 & 0.8546 & -0.43 \\
& MobileNetV3 \cite{howard2019searching} & 0.8090 & 0.9326 & 0.8071 & 0.8306 & -2.91 & 0.8179 & 0.9314 & 0.8148 & 0.8366 & -2.59 \\
& EfficientNetV2 \cite{tan2021efficientnetv2} & 0.8689 & \cellcolor[HTML]{FFF2CC}\textbf{0.9602} & 0.8533 & 0.8826 & +3.08 & 0.8736 & \cellcolor[HTML]{FFF2CC}\textbf{0.9606} & 0.8581 & 0.8867 & +2.98 \\
& MobileVIT \cite{mehta2021mobilevit} & 0.8486 & 0.9462 & 0.8478 & 0.8709 & +1.05 & 0.8545 & 0.9453 & 0.8529 & 0.8748 & +1.07 \\
& EdgeNeXt \cite{maaz2022edgenext} & 0.8642 & 0.9468 & 0.8632 & 0.8836 & +2.61 & 0.8694 & 0.9458 & 0.8677 & 0.8869 & +2.56 \\
& Vision Transformer \cite{dosovitskiy2020image} & 0.8314 & 0.9568 & 0.7891 & 0.8325 & -0.67 & 0.8407 & 0.9569 & 0.7985 & 0.8403 & -0.31 \\
& ConvNeXt \cite{liu2022convnet} & 0.8784 & 0.9446 & 0.8705 & \cellcolor[HTML]{F1E9DF}0.8891 & +4.03 & \cellcolor[HTML]{F1E9DF}0.8829 & 0.9469 & 0.8746 & \cellcolor[HTML]{F1E9DF}0.8932 & \cellcolor[HTML]{F1E9DF}+3.91 \\
\hline
\multirow{5}{*}{Pretrained Model}
& MAE \cite{he2022masked} & 0.8134 & 0.9252 & 0.7681 & 0.8069 & -2.47 & 0.8235 & 0.9292 & 0.7786 & 0.8170 & -2.03 \\
& SimMIM \cite{xie2022simmim} & 0.8706 & \cellcolor[HTML]{F1E9DF}0.9595 & 0.8553 & 0.8806 & +3.25 & 0.8730 & \cellcolor[HTML]{F1E9DF}0.9597 & 0.8567 & 0.8817 & +2.92 \\
& DINO \cite{caron2021emerging} & 0.8169 & 0.9071 & 0.7816 & 0.8135 & -2.12 & 0.8258 & 0.9120 & 0.7902 & 0.8222 & -1.80 \\
& MoCoV3 \cite{chen2021empirical} & 0.8460 & 0.9425 & 0.8447 & 0.8663 & +0.79 & 0.8539 & 0.9431 & 0.8515 & 0.8719 & +1.01 \\
& SimCLR \cite{chen2020simple} & 0.8469 & 0.9414 & 0.8456 & 0.8689 & +0.88 & 0.8520 & 0.9407 & 0.8501 & 0.8720 & +0.82 \\
\hline
\multirow{4}{*}{VLM}
& CLIP \cite{radford2021learning} & \cellcolor[HTML]{FFF2CC}\textbf{0.9291} & 0.9453 & \cellcolor[HTML]{FFF2CC}\textbf{0.9295} & \cellcolor[HTML]{FFF2CC}\textbf{0.9218} & \cellcolor[HTML]{FFF2CC}\textbf{+9.10} & \cellcolor[HTML]{FFF2CC}\textbf{0.9256} & 0.9448 & \cellcolor[HTML]{FFF2CC}\textbf{0.9251} & \cellcolor[HTML]{FFF2CC}\textbf{0.9183} & \cellcolor[HTML]{FFF2CC}\textbf{+8.18} \\
& BLIP \cite{li2022blip} & \cellcolor[HTML]{F1E9DF}0.8821 & 0.8760 & \cellcolor[HTML]{F1E9DF}0.8778 & 0.8617 & \cellcolor[HTML]{F1E9DF}+4.40 & 0.8827 & 0.8804 & \cellcolor[HTML]{F1E9DF}0.8777 & 0.8626 & +3.89 \\
& BLIP2 \cite{li2023blip} & 0.8715 & 0.9300 & 0.8678 & 0.8771 & +3.34 & 0.8772 & 0.9254 & 0.8700 & 0.8763 & +3.34 \\
& LLaVA \cite{liu2023visual} & 0.8655 & 0.9315 & 0.8600 & 0.8755 & +2.74 & 0.8690 & 0.9320 & 0.8640 & 0.8790 & +2.52 \\
\hline
\end{tabular}
\label{TableRare}
\vspace{-5pt}
\end{table*}

\begin{table}[t!]
\renewcommand\arraystretch{1.3}
\caption{\textbf{Performance on Low-Clarity Samples in the TS-1M Benchmark.} 
Results are reported on traffic sign crops with resolutions below $30 \times 30$ pixels before model-input resizing. $\Delta$ denotes the Acc.@1 relative improvement over the ResNet50 baseline. \best{} and \second{} highlight the best and second-best results, respectively.}
\centering
% \footnotesize
\scriptsize
% \begin{tabular}{>{\centering\arraybackslash}m{2.7cm}ccccc}
\begin{tabular}{lccccc}
\hline
\textbf{Model} & \textbf{Acc.@1} & \textbf{Pre.} & \textbf{Recall} & \textbf{F-1} & \textbf{$\Delta$ (\%)} \\
\hline
ResNet50 \cite{he2016deep}           & 0.8768 & 0.7707 & 0.7768 & 0.7553 & - \\
ResNet101 \cite{he2016deep}          & 0.8876 & 0.7824 & 0.8414 & 0.7948 & +1.08 \\
ResNext50 \cite{xie2017aggregated}   & 0.9055 & 0.8005 & 0.8512 & 0.8123 & +2.87 \\
ShufflenetV2 \cite{ma2018shufflenet} & 0.8691 & 0.7284 & 0.8082 & 0.7468 & -0.77 \\
MobileNetV3 \cite{howard2019searching} & 0.8789 & 0.7882 & 0.7925 & 0.7729 & +0.21 \\
EfficientNetV2 \cite{tan2021efficientnetv2} & 0.9014 & 0.8284 & 0.8325 & 0.8164 & +2.46 \\
MobileVIT \cite{mehta2021mobilevit} & 0.8845 & 0.7624 & 0.8098 & 0.7656 & +0.77 \\
EdgeNeXt \cite{maaz2022edgenext} & 0.8937 & 0.8014 & 0.8518 & 0.8121 & +1.69 \\
Vision Transformer \cite{dosovitskiy2020image} & 0.8622 & 0.8184 & 0.7655 & 0.7659 & -1.46 \\
ConvNeXt \cite{liu2022convnet} & 0.9088 & 0.8426 & \cellcolor[HTML]{F1E9DF}0.8725 & 0.8441 & +3.20 \\
\hline
MAE \cite{he2022masked} & 0.8332 & 0.7943 & 0.7585 & 0.7536 & -4.36 \\
SimMIM \cite{xie2022simmim} & 0.9000 & 0.8348 & 0.8502 & 0.8266 & +2.32 \\
DINO \cite{caron2021emerging} & 0.8394 & 0.8076 & 0.8047 & 0.7858 & -3.74 \\
MoCoV3 \cite{chen2021empirical} & 0.9012 & 0.7993 & 0.8336 & 0.7995 & +2.44 \\
SimCLR \cite{chen2020simple} & 0.8959 & 0.7885 & 0.8293 & 0.7938 & +1.91 \\
\hline
CLIP \cite{radford2021learning} & \cellcolor[HTML]{FFF2CC}\textbf{0.9313} & \cellcolor[HTML]{FFF2CC}\textbf{0.8861} & 0.8583 & \cellcolor[HTML]{FFF2CC}\textbf{0.8583} & \cellcolor[HTML]{FFF2CC}\textbf{+5.45} \\
BLIP \cite{li2022blip} & \cellcolor[HTML]{F1E9DF}0.9175 & \cellcolor[HTML]{F1E9DF}0.8491 & \cellcolor[HTML]{FFF2CC}\textbf{0.8835} & \cellcolor[HTML]{F1E9DF}0.8442 & \cellcolor[HTML]{F1E9DF}+4.07 \\
BLIP2 \cite{li2023blip} & 0.8931 & 0.7621 & 0.8214 & 0.7685 & +1.63 \\
LLAVA \cite{liu2023visual} & 0.8926 & 0.7907 & 0.7531 & 0.7526 & +1.58 \\
\hline
\end{tabular}
\label{TableClarity}
\vspace{-12pt}
\end{table}

\subsection{Robustness on Low-Clarity Signs}

Table~\ref{TableClarity} reports performance on the low-clarity set. Among classic architectures, ConvNeXt, ResNeXt50, and EfficientNetV2 achieve the strongest results, with ConvNeXt reaching 0.9088, or 3.20 percentage points above ResNet50.

Among SSL models, MoCoV3 achieves the highest low-clarity accuracy (0.9012), followed by SimMIM (0.9000) and SimCLR (0.8959). All three exceed the ResNet50 baseline, whereas MAE and DINO fall below it by 4.36 and 3.74 percentage points, respectively. VLMs obtain the best overall performance, with CLIP achieving the highest accuracy of 0.9313 (+5.45\%) and BLIP also showing strong gains. These results highlight the advantage of text-aligned semantic priors in compensating for missing visual details. All models obtain lower accuracy than on the full test set, underscoring the challenge posed by limited spatial detail.

\begin{table*}
\renewcommand\arraystretch{1.3}
% \caption{Semantic Enhancement}
\caption{\textbf{Effect of Semantic Descriptions on VLM Classification.} 
The table compares zero-shot inference, fine-tuning without semantic descriptions, and fine-tuning with semantic descriptions under different prompt compositions (class name, scenario, and rules). \best{} and \second{} highlight the best and second-best results, respectively.}
% \vspace{-8pt}
\centering
\setlength{\tabcolsep}{2.05mm}
% \footnotesize
\scriptsize
% \begin{tabular}{@{}ccccccc>{\columncolor{gray!20}}c@{}}
\begin{tabular}{l@{}ccc|cccc|cccc|cccc>{\columncolor{gray!20}}c@{}}
\hline
\multirow{2}{*}{\textbf{Model}} & \multicolumn{3}{c|}{ \textbf{Semantic Description} } & \multicolumn{4}{c|}{ \textbf{Zero-shot} } &  \multicolumn{4}{c|}{ \textbf{Finetuned w/o Description} } & \multicolumn{4}{c}{ \textbf{Finetuned w/ Description} } \\
\cline{2-16} 
 & \textbf{Class} & \textbf{Scenario} & \textbf{Rules} & \textbf{Acc.@1} & \textbf{Pre.} & \textbf{Recall} & \textbf{F-1} & \textbf{Acc.@1} & \textbf{Pre.} & \textbf{Recall} & \textbf{F-1} & \textbf{Acc.@1} & \textbf{Pre.} & \textbf{Recall} & \textbf{F-1}\\
\hline \multirow{4}{*}{CLIP \cite{radford2021learning}} 
  & \textcolor[HTML]{228B22}{\ding{51}} & \textcolor[HTML]{B22222}{\ding{55}} & \textcolor[HTML]{B22222}{\ding{55}} & 0.0664 & 0.1227 & 0.0918 & 0.0730 & \cellcolor[HTML]{FFF2CC}\textbf{0.9554} & \cellcolor[HTML]{FFF2CC}\textbf{0.9359} & \cellcolor[HTML]{FFF2CC}\textbf{0.9040} & \cellcolor[HTML]{FFF2CC}\textbf{0.9105} & 0.9426 & 0.9094 & 0.8895 & 0.8849 \\
  & \textcolor[HTML]{228B22}{\ding{51}} & \textcolor[HTML]{228B22}{\ding{51}} & \textcolor[HTML]{B22222}{\ding{55}} & 0.0805 & 0.1067 & 0.1051 & 0.0782 & 0.8149 & 0.7621 & 0.8394 & 0.7436 & 0.9514 & 0.9043 & 0.9009 & 0.8928 \\
  & \textcolor[HTML]{228B22}{\ding{51}} & \textcolor[HTML]{B22222}{\ding{55}} & \textcolor[HTML]{228B22}{\ding{51}} & 0.0851 & 0.1103 & 0.1057 & 0.0794 &  0.7450 & 0.6827 & 0.8032 & 0.6695 & 0.9530 & 0.9179 & 0.9116 & 0.9014 \\
  & \textcolor[HTML]{228B22}{\ding{51}} & \textcolor[HTML]{228B22}{\ding{51}} & \textcolor[HTML]{228B22}{\ding{51}} & 0.0971 & 0.1166 & 0.1068 & 0.0822 & 0.6739 & 0.6612 & 0.7753 & 0.6281 & \cellcolor[HTML]{F1E9DF}0.9581 & \cellcolor[HTML]{F1E9DF}0.9345 & \cellcolor[HTML]{F1E9DF}0.9161 & \cellcolor[HTML]{F1E9DF}0.9160  \\
\hline \multirow{4}{*}{BLIP \cite{li2022blip}} 
 & \textcolor[HTML]{228B22}{\ding{51}} & \textcolor[HTML]{B22222}{\ding{55}} & \textcolor[HTML]{B22222}{\ding{55}}  & 0.1432 & 0.2060 & 0.1969 & 0.1426 & \cellcolor[HTML]{F1E9DF}0.9448 & 0.8872 & \cellcolor[HTML]{F1E9DF}0.8949 & 0.8770 & 0.8958 & 0.8111 & 0.8574 & 0.8158 \\
 & \textcolor[HTML]{228B22}{\ding{51}} &  \textcolor[HTML]{228B22}{\ding{51}} & \textcolor[HTML]{B22222}{\ding{55}} & 0.1467 & \cellcolor[HTML]{F1E9DF}0.2228 & \cellcolor[HTML]{F1E9DF}0.2286 & \cellcolor[HTML]{F1E9DF}0.1652 & 0.7971 & 0.7817 & 0.8241 & 0.7396  & 0.9118 & 0.8580 & 0.8898 & 0.8261 \\
 & \textcolor[HTML]{228B22}{\ding{51}} & \textcolor[HTML]{B22222}{\ding{55}} & \textcolor[HTML]{228B22}{\ding{51}}  & 0.1522 & 0.2194 & 0.2005 & 0.1517 & 0.7287 & 0.7020 & 0.7849 & 0.6723  & 0.9280 & 0.8721 & 0.8819 & 0.8507 \\
 & \textcolor[HTML]{228B22}{\ding{51}} & \textcolor[HTML]{228B22}{\ding{51}}  & \textcolor[HTML]{228B22}{\ding{51}} & \cellcolor[HTML]{FFF2CC}\textbf{0.1754} & \cellcolor[HTML]{FFF2CC}\textbf{0.2370} & \cellcolor[HTML]{FFF2CC}\textbf{0.2453} &  \cellcolor[HTML]{FFF2CC}\textbf{0.1705} & 0.6642 & 0.7172 & 0.7579 & 0.6377  & 0.9478 & 0.8881 & 0.9098 & 0.8834 \\
\hline \multirow{4}{*}{LLAVA \cite{liu2023visual}}  
 & \textcolor[HTML]{228B22}{\ding{51}} & \textcolor[HTML]{B22222}{\ding{55}} & \textcolor[HTML]{B22222}{\ding{55}}  & 0.1061 & 0.1094 & 0.0868 & 0.0866 & 0.9406 & \cellcolor[HTML]{F1E9DF}0.9136 & 0.8677 & \cellcolor[HTML]{F1E9DF}0.8776 & 0.9416 & 0.8895 & 0.8953 & 0.8746 \\ 
 & \textcolor[HTML]{228B22}{\ding{51}} & \textcolor[HTML]{228B22}{\ding{51}} & \textcolor[HTML]{B22222}{\ding{55}}  & 0.1566 & 0.1523 & 0.1049 & 0.1194 & 0.8859 & 0.8607 & 0.8033 & 0.8104 & 0.9507 & 0.9189 & 0.8943 & 0.8971  \\
 & \textcolor[HTML]{228B22}{\ding{51}} & \textcolor[HTML]{B22222}{\ding{55}} & \textcolor[HTML]{228B22}{\ding{51}}  & 0.1460 & 0.1452 & 0.1176 & 0.1099 & 0.8685 & 0.8307 & 0.7743 & 0.7819  & 0.9452 & 0.9244 & 0.9106 & 0.9065 \\
 & \textcolor[HTML]{228B22}{\ding{51}} & \textcolor[HTML]{228B22}{\ding{51}} & \textcolor[HTML]{228B22}{\ding{51}}  & \cellcolor[HTML]{F1E9DF}0.1716 & 0.1500 & 0.1263 & 0.1302 & 0.8373 & 0.8106 & 0.7293 & 0.7261  & \cellcolor[HTML]{FFF2CC}\textbf{0.9698}  & \cellcolor[HTML]{FFF2CC}\textbf{0.9493} & \cellcolor[HTML]{FFF2CC}\textbf{0.9270} & \cellcolor[HTML]{FFF2CC}\textbf{0.9337} \\
\hline
\end{tabular}
\vspace{-10pt}
\label{ablation}
% \vspace{-10pt}
\end{table*}

\subsection{Effect of Semantic Enhancement}

Table~\ref{ablation} evaluates the effect of semantic descriptions on classification by CLIP, BLIP, and LLaVA. The text components are the category name (\textit{Class}), scene context (\textit{Scenario}), and traffic-rule information (\textit{Rules}). We compare zero-shot inference and models fine-tuned with or without descriptions. In description-based fine-tuning, Scenario and Rules are incorporated into the training prompts. For zero-shot models, these descriptions are instead supplied through inference prompts without parameter updates. The four text compositions specify the inference prompts for each setting. The predefined category vocabulary remains unchanged, and test-image ground-truth labels are neither provided as inputs nor used to select descriptions.

For zero-shot models, adding both scenario and rule descriptions improves accuracy and macro-F1 over class-only inputs for all three VLMs. Absolute recognition accuracy nevertheless remains low.

For models fine-tuned without descriptions, adding scenario or rule descriptions at inference reduces classification accuracy, with smaller declines for LLaVA than for CLIP and BLIP. This pattern highlights sensitivity to changes in textual inputs between training and inference.

With description-based fine-tuning, the Class + Scenario + Rules configuration achieves the highest accuracy for each model, reaching 0.9581 for CLIP, 0.9478 for BLIP, and 0.9698 for LLaVA. These gains measure the classification benefit of semantic descriptions, rather than directly testing traffic-rule understanding.

\subsection{Visualization and Qualitative Analysis}

\begin{figure*}[t!]
    \centering
    \includegraphics[width=1.0\textwidth]{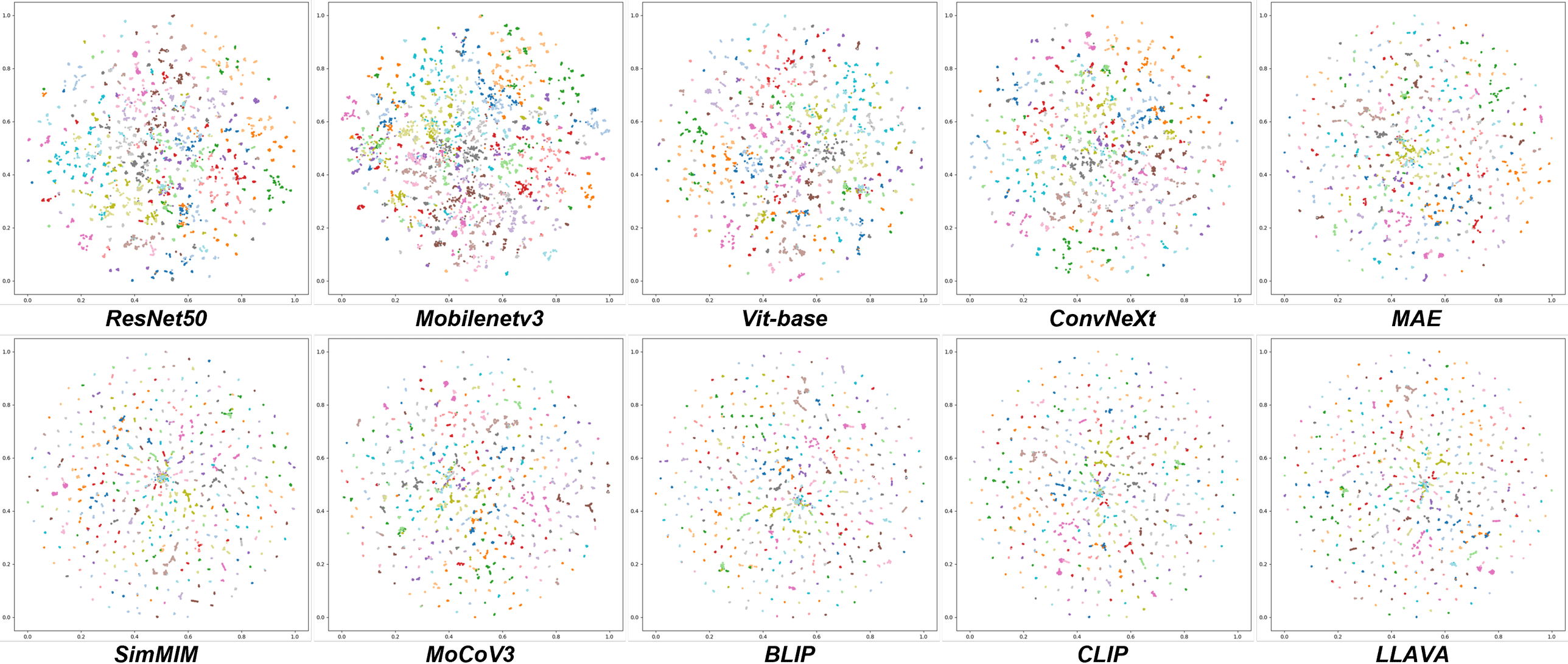}
    \vspace{-12pt}
    \caption{\textbf{t-SNE visualization of feature representations learned by different models on TS-1M.} Classic supervised models exhibit scattered and overlapping clusters, while self-supervised models achieve improved compactness. VLMs form the most structured and semantically coherent clusters, highlighting the benefit of image-text alignment for TSR.}
    \label{vis-tsne}
    % \vspace{-5pt}
\end{figure*}

\begin{figure*}[t!]
    \centering
    \includegraphics[width=1.0\textwidth]{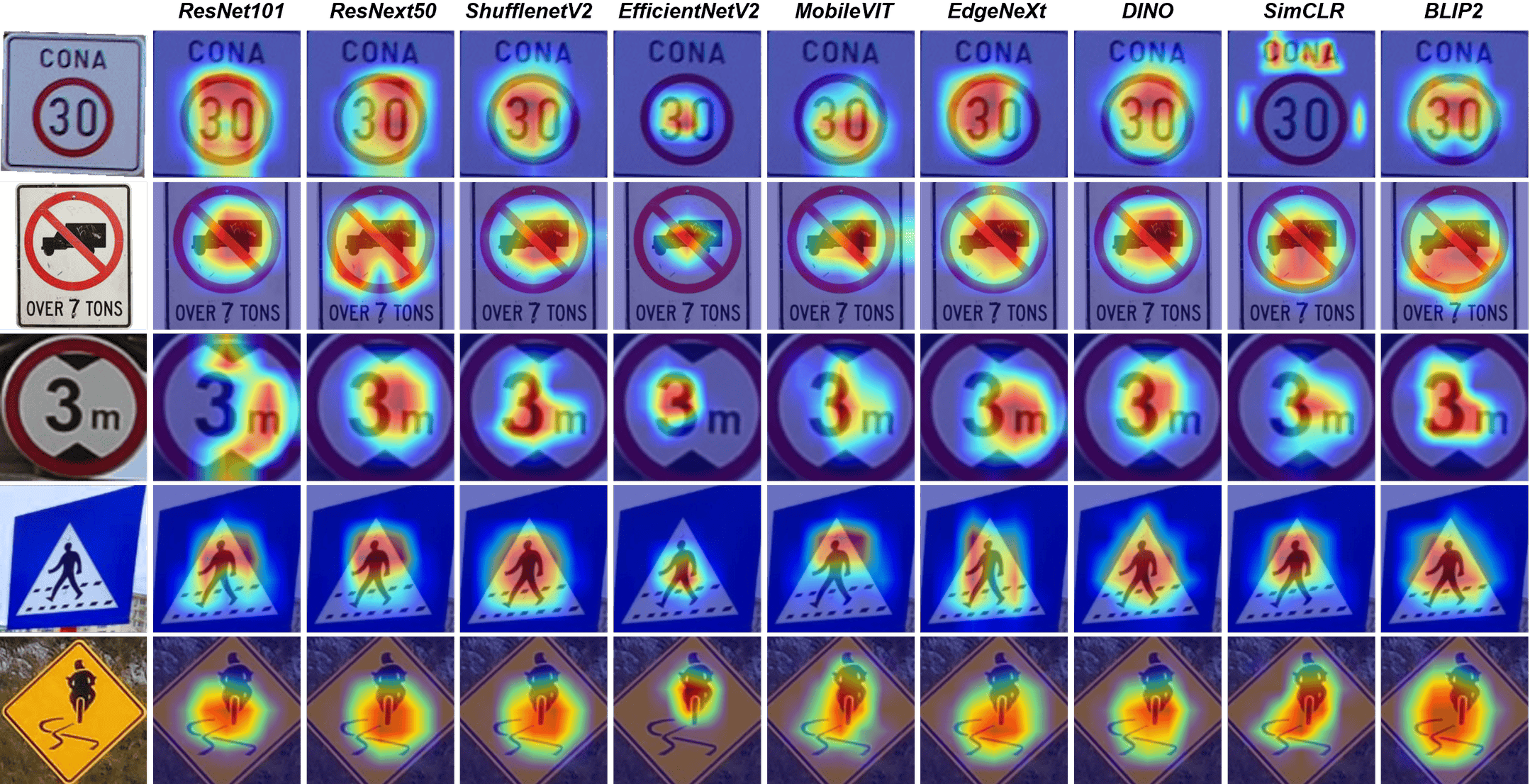}
    \vspace{-10pt}
    \caption{\textbf{Grad-CAM visualization of model attention on traffic sign.} 
The examples illustrate differences in highlighted sign regions across the evaluated models.}
    \label{vis-cam}
    \vspace{-12pt}
\end{figure*}

To further examine the feature representation capability of different models, we conduct two types of qualitative analyses: 2D t-SNE visualization on the test-set embeddings and Grad-CAM visualization on representative samples, as shown in Fig.~\ref{vis-tsne} and Fig.~\ref{vis-cam}. The t-SNE results reveal clear differences in the structure of the learned feature space. Classic CNNs such as ResNet50 and MobileNetV3 exhibit relatively scattered and overlapping clusters, indicating limited separation between visually similar categories. Vision Transformer and MAE show even more pronounced overlap, suggesting weaker discriminative power under the TS-1M taxonomy. In contrast, modern CNNs (ConvNeXt) and SSL models (SimMIM, MoCoV3) produce tighter intra-class clusters with improved inter-class margins. VLMs demonstrate the strongest semantic organization, with CLIP showing the most distinct and well-formed clusters, while BLIP and LLAVA also exhibit clear semantic grouping, highlighting the effectiveness of image-text alignment in capturing high-level category relations.

Grad-CAM visualization provides additional insight into the spatial focus of these models. Strong CNN baselines such as ResNet101, ResNeXt50, and EfficientNetV2 consistently highlight the most informative regions of traffic signs, including digits, shapes, and symbolic contours. In contrast, lightweight models (ShuffleNetV2, MobileVIT) show less stable attention patterns, often focusing on incomplete or peripheral regions. Among SSL models, DINO and SimCLR display broader spatial coverage in these examples. VLMs such as BLIP2 attend to more complete semantic regions. Overall, these visualizations complement the quantitative results by illustrating differences in feature clustering and spatial responses among the evaluated models.

\subsection{Time Consumption}

To provide a comprehensive evaluation of practical usability on TS-1M, we further benchmark the training and inference time of all models (Table~\ref{TableTimeConsumption}). Training efficiency is reported as the average time per epoch (h/epoch) measured on $4 \times$ A100 GPUs, while inference efficiency is evaluated in frames per second (FPS) on a single RTX 5090 GPU. This analysis complements accuracy-based comparisons by offering essential insights into computational cost, which is particularly relevant for real-world TSR systems and downstream deployment.

The results reveal a clear efficiency-capacity trade-off across different model families. Classic CNNs and lightweight architectures (e.g., ShuffleNetV2 and MobileNetV3) exhibit excellent training efficiency and high inference throughput, making them well suited for real-time or resource-constrained scenarios. Transformer-based and self-supervised pretrained models (e.g., ViT, MAE, SimMIM, and DINO) incur substantially higher training costs on TS-1M, yet maintain moderate inference speed, favoring offline training with high-performance deployment. In contrast, VLMs show orders-of-magnitude differences in computational overhead: CLIP achieves a relatively balanced trade-off between efficiency and performance. BLIP incurs cross-attention costs during ITM reranking, while BLIP2 and LLaVA additionally incur the cost of large language model inference. Overall, these results highlight that model selection should be guided by a joint consideration of recognition accuracy, training cost, and inference efficiency, and TS-1M serves as an effective benchmark for systematically analyzing this trade-off.

\subsection{Real Scene Experiment}
\begin{figure*}[t!]
    \centering
    \includegraphics[width=1.0\textwidth]{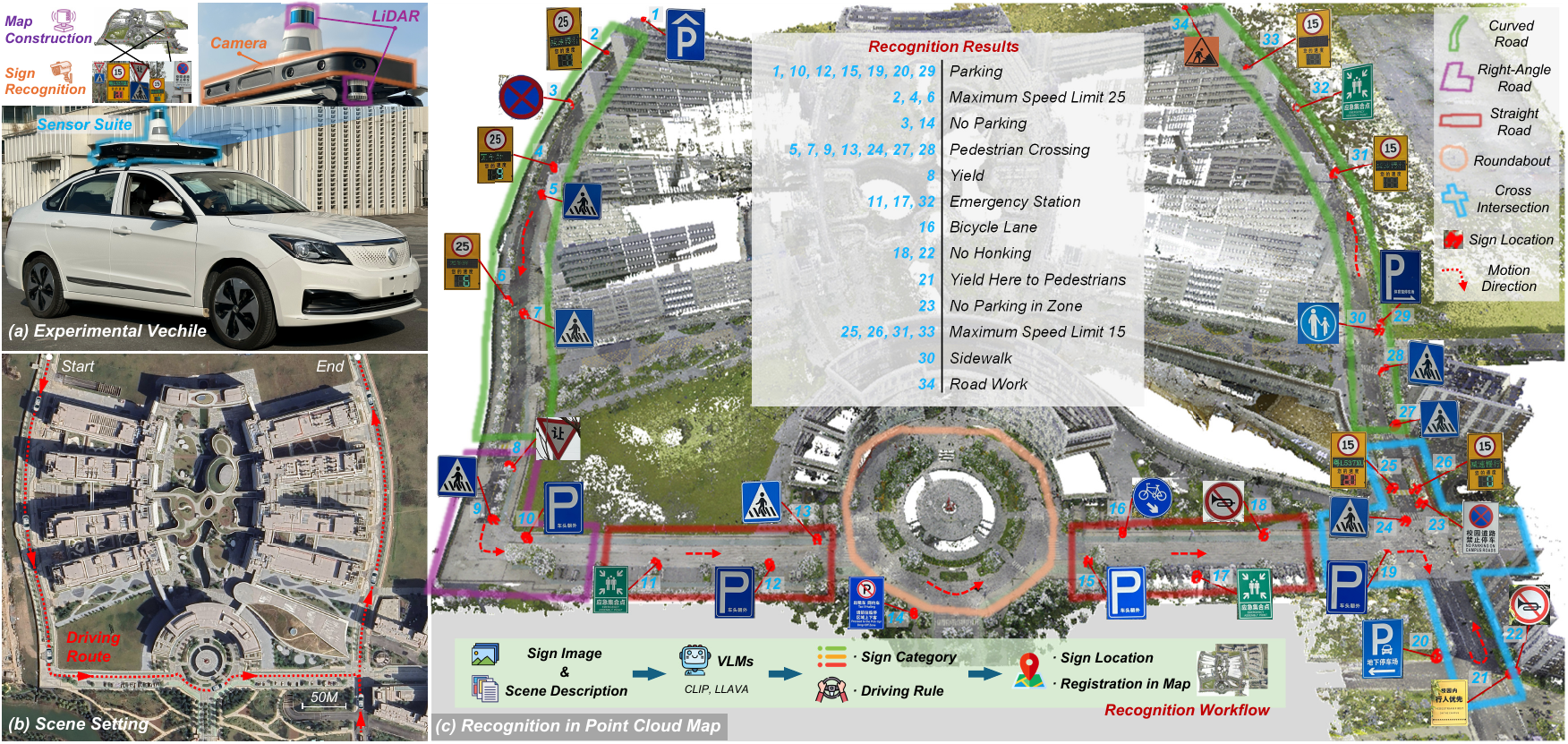}
    \vspace{-12pt}
\caption{\textbf{On-road semantic mapping with traffic sign recognition.}
(a) Autonomous vehicle with multi-sensor setup for map construction and traffic sign perception.
(b) Driving route and scene settings with diverse road structures.
(c) Traffic sign registration in the point cloud map, associating predicted categories and generated rule texts with estimated spatial locations.}
    \label{real-world}
    \vspace{-10pt}
\end{figure*}

Semantic mapping enriches geometric representations with semantic and instance information~\cite{chen2025semanticoccupancy}. We conduct an on-road semantic-mapping case study to examine how TS-1M-based traffic sign recognition can enrich a LiDAR point cloud map (Fig.~\ref{real-world}). Associating sign categories and generated rule texts with mapped locations adds an interpretable semantic layer for road-scene understanding.

The data are collected using an autonomous test vehicle equipped with a sensor suite consisting of a 128-beam high-resolution LiDAR and a surround-view camera system. The experiment is conducted on real-world university roads, covering a driving route of approximately $1.4\,\mathrm{km}$ with diverse road structures, including curved roads, right-angle turns, straight segments, roundabouts, and cross intersections. LiDAR data are utilized to construct a high-definition point cloud map, while the surround-view camera system is responsible for visual traffic sign perception and recognition.

A CLIP model fine-tuned on TS-1M predicts the category of each detected traffic sign crop. LLaVA processes the sign image and a description of the surrounding road scene to generate traffic-rule text. Camera-LiDAR registration associates each sign with a three-dimensional map position, as detailed in the Supplementary Material.

The 34 registered instances are represented as map entries linking the predicted sign category, estimated three-dimensional position, and generated rule text. This connection makes recognition outputs spatially interpretable within the mapped road scene.

\begin{table}[t!]
\renewcommand\arraystretch{1.2}
\caption{\textbf{Time Consumption on the TS-1M Dataset.} 
Training time per epoch and inference FPS are reported for all evaluated models. \best{} and \second{} highlight the best and second-best results, respectively.}
\centering
% \footnotesize
\scriptsize
% \begin{tabular}{>{\centering\arraybackslash}m{2.7cm}cc}
\begin{tabular}{lccc}
\hline
\textbf{Model} & \begin{tabular}[c]{@{}c@{}}\textbf{Training (h/epoch) $\downarrow$}\\ \textbf{4 × A100 GPU}\end{tabular} & \begin{tabular}[c]{@{}c@{}}\textbf{Testing (FPS) $\uparrow$}\\ \textbf{1 × 5090 GPU}\end{tabular} \\
\hline
ResNet50 \cite{he2016deep} & \cellcolor[HTML]{F1E9DF}0.09 & \cellcolor[HTML]{F1E9DF}434  \\
ResNet101 \cite{he2016deep}          & 0.12 & 218  \\
ResNext50 \cite{xie2017aggregated}   & 0.11 & 393  \\
ShufflenetV2 \cite{ma2018shufflenet} & \cellcolor[HTML]{FFF2CC}\textbf{0.03} & 411  \\
MobileNetV3 \cite{howard2019searching} & \cellcolor[HTML]{FFF2CC}\textbf{0.03} & \cellcolor[HTML]{FFF2CC}\textbf{446}  \\
EfficientNetV2 \cite{tan2021efficientnetv2} & 0.13 & 162  \\
MobileVIT \cite{mehta2021mobilevit} & 0.15 & 258 \\
EdgeNeXt \cite{maaz2022edgenext} & 0.18 & 335  \\
Vision Transformer \cite{dosovitskiy2020image} & 0.42 & 266  \\
ConvNeXt \cite{liu2022convnet} & 0.39 & 254 \\
\hline
MAE \cite{he2022masked} & 0.41 & 272  \\
SimMIM \cite{xie2022simmim} & 0.39 & 271 \\
DINO \cite{caron2021emerging} & 0.44 & 268 \\
MoCoV3 \cite{chen2021empirical} & 0.12 & 404 \\
SimCLR \cite{chen2020simple} & 0.14 & 411  \\
\hline
CLIP \cite{radford2021learning} & 0.15 & 294  \\
BLIP \cite{li2022blip} & 5.97 & 122 \\
BLIP2 \cite{li2023blip} & 2.03 & 4 \\
LLAVA \cite{liu2023visual} & 11.89 & 8 \\
\hline
\end{tabular}
\label{TableTimeConsumption}
\vspace{-10pt}
\end{table}

\subsection{Summary of Key Findings}

The benchmark reveals substantial performance differences across models and evaluation settings. Under target-specific exclusions, CLIP achieves the highest transfer accuracy on nine of the ten targets, while modern CNNs remain competitive and self-supervised models show target-dependent performance. Together with the in-domain challenge evaluations, these results characterize the strengths and limitations of the tested model configurations under distribution shift and data imbalance.

Further analysis under challenge-oriented benchmarks leads to four key takeaways:
\textbf{(1) Target holdout enables evaluation of cross-region and cross-dataset transfer.}
With target datasets and their corresponding regions excluded from downstream training, CLIP achieves the highest accuracy on nine of the ten targets. MTSD provides a complementary evaluation on a held-out multi-region source dataset. The results characterize transfer under the specified exclusions.
\textbf{(2) Long-tailed recognition is governed by semantic-visual mismatch.}
Performance degradation in rare categories is not only due to data scarcity, but also to insufficient semantic separability, where semantic priors play a critical role in reducing uncertainty under limited visual evidence.
\textbf{(3) Low-clarity recognition exposes an information bottleneck.}
All models suffer substantial performance drops under severe degradation, indicating that semantic alignment can only partially compensate for the loss of visual information.
\textbf{(4) Semantic enhancement is not inherently beneficial.}
Descriptions can improve classification when used consistently during fine-tuning and evaluation, while changes in textual inputs can reduce recognition accuracy.

These findings suggest that future progress in TSR requires integrating visual discrimination with structured semantic alignment, while explicitly addressing distribution shift, data imbalance, and degraded visual conditions.

\section{Discussion of VLMs in TSR Challenges}
\label{sec:discussion}

The evaluation on TS-1M shows strong performance from CLIP across the overall benchmark and multiple challenge settings. In the cross-region and cross-dataset experiments, it achieves the highest accuracy on nine of the ten held-out targets. This finding is consistent with a benefit from image-text alignment for transfer across traffic sign domains. However, differences in architecture, pretraining data, and adaptation also affect the comparison, so the results do not establish semantic alignment as the sole cause of the observed advantage.

The semantic-description experiments show that classification performance depends on the consistency of textual inputs between training and evaluation. Descriptions used during fine-tuning can improve recognition, whereas introducing richer descriptions only at inference can reduce the accuracy of models fine-tuned with class names alone.

Different VLM architectures exhibit distinct trade-offs. Dual-encoder models such as CLIP provide stable performance and strong scalability, making them suitable for large-scale and cross-region TSR scenarios. In contrast, fusion-based or generative models (e.g., BLIP and LLaVA) use cross-attention or autoregressive generation, but incur higher computational cost and reduced deployment efficiency. Overall, VLMs should be viewed as a complementary paradigm that enhances TSR through semantic grounding, suggesting that future research should focus on stable vision-language alignment, principled semantic modeling, and architecture-aware deployment strategies.

\section{Conclusion}
\label{sec:conclusion}

This paper presents TS-1M, a large-scale and globally diverse traffic sign dataset together with a unified diagnostic benchmark for TSR. By addressing limitations in dataset scale, geographic coverage, and taxonomy consistency, TS-1M establishes a standardized foundation for systematic and realistic evaluation. 
Built upon this dataset, we benchmark classical supervised models, self-supervised pretrained models, and multimodal VLMs under four challenge-oriented settings: cross-region and cross-dataset generalization, rare-category recognition, low-clarity robustness, and recognition with semantic descriptions. Transfer evaluation excludes each target dataset and, for geographically localized targets, its region from downstream training. Modern CNNs remain competitive, self-supervised transfer performance varies across targets, and CLIP achieves the highest accuracy on nine of the ten held-out targets. 
Beyond offline evaluation, an on-road case study illustrates the integration of TS-1M-based recognition into semantic-map construction, linking sign categories, generated rule texts, and spatial locations.

Overall, TS-1M provides a unified diagnostic benchmark for principled TSR evaluation, enabling systematic analysis of model capabilities and limitations under realistic conditions and supporting the development of robust and scalable recognition systems.

\textbf{Limitations and Future Work.}
The current benchmark focuses on image-level classification. Extending it to detection, temporal modeling, and system-level evaluation is a promising direction. Further improving the efficiency and robustness of multimodal models is also worth future exploration.

% \newpage

{
\bibliographystyle{IEEEtran}
\bibliography{ref}
}

\vfill

\clearpage


\section*{Supplementary material}

# Traffic Sign Recognition in Autonomous Driving: Dataset, Benchmark, and Field Experiment -Supplementary Material-

Guoyang Zhao*, Weiqing Qi*, Kai Zhang, Chenguang Zhang, Zeying Gong, Zhihai Bi, Kai Chen, Benshan Ma, Ming Liu, *Fellow, IEEE*, and Jun Ma, *Senior Member, IEEE*

*Abstract*—This supplementary material provides comprehensive technical details supporting the TS-1M dataset and its diagnostic benchmark for traffic sign recognition. We first detail the construction of TS-1M, including the unified traffic sign taxonomy and data sourcing and licensing policies. We then describe standardized benchmark protocols spanning classical supervised models, self-supervised learning approaches, and vision-language models, together with unified training and evaluation settings for fair comparison. Furthermore, we present detailed learning curve analyses to characterize the optimization and generalization behavior of different model families on TS-1M. Finally, we introduce the real-world validation system, including the hardware platform, LiDAR-based mapping, and cross-modal traffic sign registration pipeline used in field experiments.

## I. TS-1M DATASET DETAILS

This section provides additional information on the construction of the TS-1M dataset that is not included in the main paper due to space limitations. We focus on two key aspects: the formation of the unified traffic sign taxonomy and the data sourcing and licensing policy.

### A. Class Taxonomy and Category Construction

The 454 traffic sign categories in TS-1M are derived through a systematic aggregation and consolidation of category definitions from multiple publicly available traffic sign datasets. Since existing datasets are typically developed for specific countries or regions, their labeling standards and category definitions differ considerably.

To construct a unified taxonomy, we first collected the complete category lists from these datasets and compiled them into a candidate category pool. The candidate categories are then manually reviewed to identify semantically equivalent labels and visually redundant classes across datasets. Categories representing the same traffic rule or visual concept are merged, while region-specific signs with distinct meanings are preserved as independent classes.

Through this consolidation process, we obtained a unified label space consisting of 454 fine-grained traffic sign categories. This taxonomy integrates the major traffic sign types appearing in existing datasets while retaining sufficient granularity to represent regional variations in traffic sign design.

### B. Data Sources and Licensing Policy

The images in TS-1M are collected from multiple publicly accessible sources, including previously published academic datasets, government open-data platforms, and curated public Internet repositories. All data sources are selected to ensure that the images are permitted for academic use and redistribution under their respective licensing terms.

During dataset construction, we conducted manual verification and filtering to remove samples with incorrect annotations, ambiguous labels, or unreliable provenance. This procedure ensures that the resulting dataset maintains consistent labeling quality and reliable data sources.

To support open and reproducible research, the TS-1M dataset will be publicly released together with all associated challenge-oriented evaluation sets.

## II. BENCHMARK TRAINING DETAILS

### A. General Experimental Settings

All benchmark models are trained on four NVIDIA A100-PCIE-40GB GPUs. Unless otherwise specified, model inference is performed on a single NVIDIA RTX 5090 GPU.

For fair comparison across different model families, we adopt a unified evaluation protocol. Specifically, we report Top-1 accuracy, macro-precision, macro-recall, and macro-F1 score to account for potential class imbalance in the TS-1M dataset.

A unified data preprocessing pipeline is applied across supervised and self-supervised models. During training, images undergo Random Resized Crop (target size $224 \times 224$) and Horizontal Flipping with a probability of 0.5. During evaluation, images are resized so that the shorter edge is 256 pixels followed by a $224 \times 224$ Center Crop. Data loading uses 10 workers per GPU with persistent workers and pinned memory enabled for efficient throughput.

Guoyang Zhao, Weiqing Qi, Kai Zhang, Zeying Gong, Zhihai Bi, Kai Chen, and Benshan Ma are with The Hong Kong University of Science and Technology (Guangzhou), Guangzhou 511453, China (e-mail: gzhao492@connect.hkust-gz.edu.cn).

Chenguang Zhang is with Lingnan University, Hong Kong SAR, China (e-mail: qwe934063437@gmail.com).

Ming Liu is with Shenzhen Unity Drive Innovation Technology Co., Ltd., Shenzhen 518063, China (e-mail: liu.ming.prc@gmail.com).

Jun Ma is with The Hong Kong University of Science and Technology (Guangzhou), Guangzhou 511453, China, and also with The Hong Kong University of Science and Technology, Hong Kong SAR, China (e-mail: jun.ma@ust.hk).

* indicates equal contribution.

## B. Classical Supervised Models Fine-Tuning

We evaluate ten representative supervised architectures covering convolutional, transformer, and hybrid designs. CNN-based models include ResNet-50 [1], ResNet-101, ResNeXt-50 [2], ShuffleNetV2 [3], MobileNetV3 [4], EfficientNetV2 [5], and ConvNeXt [6]. The pure transformer architecture is represented by Vision Transformer (ViT) [7], while hybrid CNN-transformer designs include MobileViT [8] and EdgeNeXt [9]. All experiments are implemented using the OpenMMLab `mmpretrain` framework [10].

All models are initialized with official ImageNet pre-trained weights. The original classification heads are replaced with randomly initialized linear classifiers (`LinearClsHead`) whose output dimension matches the 454 categories of TS-1M. Training is conducted for 40 epochs with a global batch size of 128 per GPU.

For classical CNN architectures such as the ResNet family, optimization is performed using SGD with a base learning rate of 0.1, momentum 0.9, and weight decay $1 \times 10^{-4}$. The learning rate follows a MultiStepLR schedule with decay at epochs 10, 20, and 30.

For transformer-based and modern hybrid architectures such as EdgeNeXt, we adopt AdamW optimization with an initial learning rate of 0.006, betas $(0.9, 0.999)$, and weight decay 0.05. Parameter-wise configurations disable weight decay for biases, normalization layers, and positional embeddings. Gradient clipping with a maximum norm of 5.0 is applied. Training uses a 3-epoch linear warmup followed by cosine annealing with a minimum learning rate of $1 \times 10^{-6}$. An exponential moving average (EMA) hook with momentum $4 \times 10^{-5}$ is further employed to stabilize training.

## C. Self-Supervised Learning Models Fine-Tuning

To complement supervised baselines, we evaluate several representative self-supervised learning (SSL) methods covering different representation paradigms. Contrastive learning models include SimCLR [11] and MoCoV3 [12] with ResNet-50 backbones. Self-distillation is represented by DINO [13] using a ViT-B/16 backbone. Masked image modeling approaches include MAE [14] (ViT-B/16) and SimMIM [15] (Swin-Base/16).

All SSL experiments are implemented using the `mmpretrain` framework. Instead of relying on standard label-dependent supervised ImageNet weights, each model is initialized with its official self-supervised checkpoint pre-trained on the ImageNet-1K dataset. The original projection heads are replaced with randomly initialized classification heads whose output dimension matches the 454 categories of TS-1M.

During fine-tuning, both the pretrained encoder and the classification head are jointly optimized for 30 epochs with a batch size of 128 per GPU. We retain the architecture-specific fine-tuning configurations provided in `mmpretrain` to ensure fair comparison.

For ResNet-based contrastive models (e.g., SimCLR and MoCoV3), optimization uses SGD with a learning rate of 0.1, momentum 0.9, and weight decay $1 \times 10^{-4}$. The learning rate follows a MultiStepLR schedule with decay at epochs 10 and 20.

For transformer-based SSL models (e.g., MAE, DINO, and SimMIM), optimization uses AdamW with a base learning rate of $6 \times 10^{-4}$ and weight decay 0.3. Gradient clipping with maximum norm 1.0 is applied, and weight decay is disabled for positional embeddings and the `[CLS]` token. The learning rate schedule consists of a 3-epoch linear warmup followed by cosine annealing. Label smoothing with value 0.1 is further applied to improve generalization.

## D. Vision Language Models (VLMs) Fine-Tuning

*1) CLIP:* CLIP experiments are conducted using the `OpenCLIP` framework [16]. The visual encoder adopts the ViT-B/16 architecture initialized with the official OpenAI CLIP pre-trained weights [17].

The model is fine-tuned on TS-1M for 20 epochs using the default OpenCLIP optimization configuration. To preserve multimodal alignment, classification is implemented through the standard prompt-based inference mechanism rather than introducing a task-specific classifier. For each category, a text embedding is generated using the template *"a photo of a {label}."*. During inference, image and text embeddings are L2-normalized and cosine similarity is computed between image features and all candidate text descriptions. The predicted label corresponds to the highest similarity score.

*2) BLIP:* BLIP experiments use the official BLIP implementation [18]. The model is initialized with weights pretrained for image-text retrieval on COCO and fine-tuned on TS-1M for 20 epochs.

Direct evaluation of Image-Text Matching (ITM) across the entire test set would require exhaustive cross-attention between every image and category description. To reduce computational cost, we adopt a two-stage candidate generation and re-ranking strategy.

First, a semantic similarity matrix between class names is computed using the sentence embedding model `Qwen3-Embedding-0.6B` [19]. For each test sample, a candidate pool is formed from the ground-truth class and its most semantically similar neighbors, resulting in a candidate set size of $K = 50$. The fine-tuned BLIP model then computes ITM probabilities only within this reduced candidate set, and the class with the highest probability is selected as the prediction.

*3) BLIP2:* BLIP-2 experiments follow the VLMClassifier fine-tuning protocol [20] using the official BLIP-2 implementation [21]. We adopt the `caption_coco_opt2.7b` configuration combining a ViT encoder with the OPT-2.7B language model.

To reduce computational cost, the ViT encoder remains frozen. Traffic sign recognition is formulated as a conditional image captioning task. Images are resized to $364 \times 364$, and the model generates the sign category using the prefix *"a photo of "*. Fine-tuning runs for 20 epochs with global batch size 32 using AMP. Optimization follows a linear warmup cosine schedule with learning rate increasing from $1 \times 10^{-8}$ to $2 \times 10^{-5}$ during the first 1000 steps and gradually decaying

thereafter. Weight decay is set to 0.05. Predictions are obtained via greedy decoding and evaluated using exact string matching against ground-truth class names.

*4) LLaVA:* LLaVA experiments follow the VLMClassifier protocol using the official LLaVA framework [22]. We adopt the `LLaVA-v1.5-7B` architecture combining a CLIP ViT-L/14 vision encoder with a large language model operating at $336 \times 336$ resolution.

During fine-tuning, both the vision encoder and the language model remain frozen while only the multimodal projector is optimized. The projector is implemented as a two-layer MLP with GELU activation. Traffic sign recognition is formulated as a visual question answering task using the prompt *"What traffic sign is shown in the photo? Reply with the traffic sign class only."*

The projector is trained for 20 epochs with batch size 32 using cosine learning rate scheduling with 3% warmup. The peak learning rate is $2 \times 10^{-4}$ without weight decay. Distributed training uses DeepSpeed ZeRO-2 together with bfloat16 mixed precision and gradient checkpointing. During inference, the generated textual responses are matched against ground-truth class names to compute classification metrics.

## III. MODEL LOSS AND VALIDATION CURVES

### A. Learning Curves on TS-1M

To analyze the optimization behavior of different model families, we visualize the training loss and validation Top-1 accuracy curves of all evaluated models during fine-tuning on TS-1M. As shown in Fig. 1, most models exhibit stable convergence, indicating that the adopted training protocol is effective across architectures and pretraining paradigms.

Classical CNN-based models, including ResNet, Efficient-NetV2, and ConvNeXt, show a rapid decrease in training loss during early epochs, accompanied by a steep increase in validation accuracy. Hybrid architectures such as MobileViT and EdgeNeXt follow a similar trend and typically reach performance saturation within 10-20 epochs. These results confirm the effectiveness of convolutional inductive biases for traffic sign recognition.

Self-supervised pretrained models also demonstrate efficient adaptation. Models such as SimMIM, MoCoV3, and Sim-CLR exhibit relatively smooth optimization trajectories and competitive validation performance. In contrast, the standard ViT baseline shows more noticeable fluctuations in validation accuracy despite decreasing training loss, suggesting that pure Transformer architectures may be less stable for fine-grained visual recognition without strong locality priors.

VLMs (CLIP, BLIP, BLIP-2, and LLaVA) exhibit the smoothest optimization behavior. Their training loss decreases steadily and validation accuracy improves consistently with minimal oscillation, indicating that multimodal pretraining provides favorable initialization for downstream TSR.

### B. Convergence and Overfitting Analysis

The training curves further reveal differences in convergence speed and generalization behavior across model families. CNN-based and hybrid architectures converge relatively early and maintain stable validation accuracy after saturation, suggesting efficient optimization with limited overfitting under the adopted training schedule.

For self-supervised models, convergence behavior is more heterogeneous. Contrastive and hybrid SSL approaches, such as MoCoV3 and SimCLR, show stable training dynamics, while reconstruction-based models like MAE and the vanilla ViT exhibit stronger fluctuations after initial convergence.

VLMs display the most stable training dynamics overall. Their loss curves remain smooth throughout fine-tuning and validation accuracy increases steadily without noticeable late-stage degradation. Despite their larger parameter scales, none of the evaluated VLMs shows clear signs of severe overfitting. This behavior suggests that multimodal pretraining provides implicit regularization that improves both optimization stability and generalization.

## IV. REAL-WORLD VALIDATION AND SYSTEM DETAILS

To evaluate real-world performance, we conduct on-road experiments at the university campus using an autonomous test vehicle equipped with multi-modal perception sensors, including 3D LiDARs and surround-view RGB cameras. As shown in Fig. 2, the experimental route spans approximately $1.4\,\mathrm{km}$ and covers multiple road structures such as straight segments, curved roads, and intersections. Along this route, 34 traffic signs are detected and registered into the point cloud map with their semantic categories and spatial positions.

### A. Hardware Setup

The hardware configuration is illustrated on the left side of Fig. 2. The platform is equipped with a RoboSense Ruby Plus 128-line LiDAR mounted at the highest position, providing a vertical field of view from $-25°$ to $+15°$. Two Ouster OS0 128-line LiDARs are installed symmetrically below the sensor deck with a vertical FOV of $-45°$ to $+45°$, providing dense 3D observations for environment perception and mapping.

For visual sensing, eight Basler acA2440-20gc cameras provide surround-view coverage. Four forward-facing cameras include two $8\,\mathrm{mm}$ lenses and two $12\,\mathrm{mm}$ lenses, while the front-left and front-right cameras use $8\,\mathrm{mm}$ lenses and the rear cameras use $4\,\mathrm{mm}$ lenses for wider coverage. For precise localization, the platform integrates a PwrPak7/7D-E1 navigation system. All perception and processing modules run on a custom onboard computer with Ubuntu 20.04, equipped with a 13th-Gen Intel Core i9 processor, an NVIDIA RTX 3080 Ti GPU, and 64 GB DDR5 memory.

### B. Point Cloud Map Construction and Sign Registration

The map construction and sign registration pipeline is illustrated in Fig. 2(a)-(d). We employ a LiDAR-inertial SLAM framework to estimate vehicle ego-pose and incrementally build a local point cloud map by integrating LiDAR and IMU measurements. The local map is subsequently aligned to a pre-built dense campus-scale point cloud map using Iterative Closest Point (ICP), producing a globally consistent map (Fig. 2(a)).

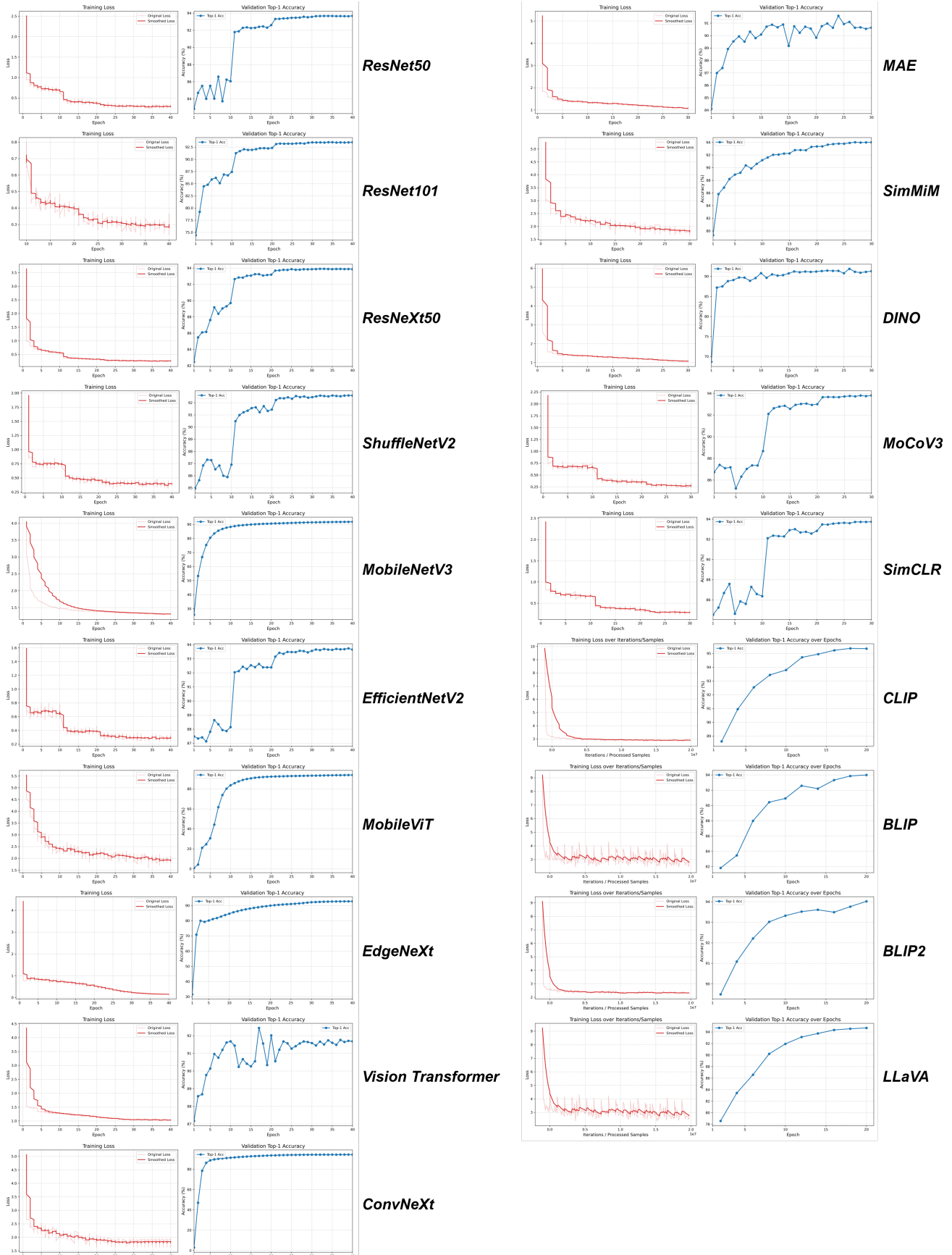

Fig. 1: **Training loss and validation Top-1 accuracy curves of different models on TS-1M.** The figure compares the optimization dynamics of classical supervised models, self-supervised pretrained models, and VLMs during fine-tuning.

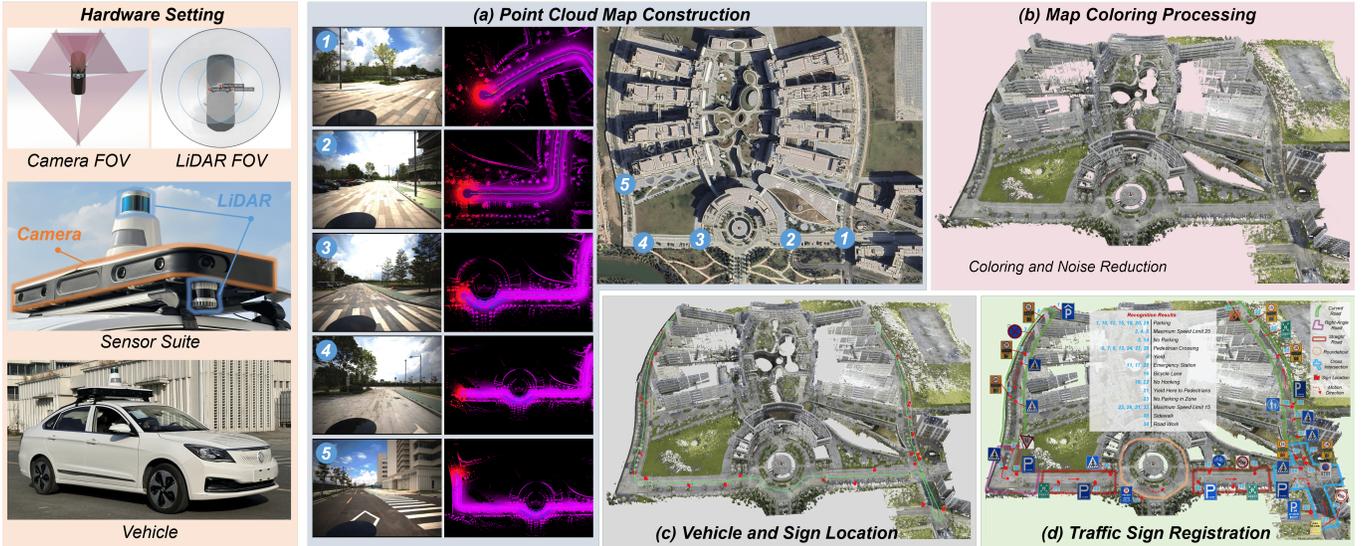

Fig. 2: **Real-world autonomous driving system setup and map construction pipeline.** Left: hardware configuration of the vehicle platform. (a) LiDAR-based point cloud map construction. (b) Map coloring and noise reduction. (c) Vehicle trajectory and sign locations. (d) Traffic sign registration with semantic categories and positions.

After registration, map coloring and noise reduction are applied to improve visualization and structural clarity, as shown in Fig. 2(b). The vehicle trajectory and preliminary traffic sign detections are then overlaid onto the map (Fig. 2(c)).

To register traffic signs into the global map, we adopt a cross-modal projection pipeline. After camera-LiDAR extrinsic calibration, the centers of detected 2D bounding boxes are projected into the point cloud via ray casting to estimate their corresponding 3D locations. Points outside the projected field of view are filtered, and the remaining points are clustered to estimate the geometric centroid of each sign. The resulting local coordinates are then transformed into the global frame using the estimated vehicle pose. As illustrated in Fig. 2(d), this process enables the construction of a semantically annotated traffic sign map, where each sign is associated with both its semantic category and spatial location.

### C. Recognition Results

Following the real-scene perception pipeline described in the main paper, all traffic signs observed along the experimental route are processed through the proposed semantic recognition framework. Specifically, a CLIP model fine-tuned on TS-1M first performs traffic sign category recognition, and the recognized sign image together with a brief scene description is then provided to the LLaVA model to generate corresponding driving rules through question-answer based reasoning.

Table I presents representative examples of the recognition and semantic reasoning results obtained during the real-world experiment. For each detected traffic sign, the system outputs the recognized category and a corresponding semantic rule that describes its driving implication within the current road context. These results demonstrate that the TS-1M trained recognition model can reliably identify traffic sign categories in real scenes, while the VLMs-based reasoning module can further transform these visual observations into structured

traffic rules that are directly usable for downstream planning and decision-making.

TABLE I: Traffic sign recognition and semantic rule generation in real-scene experiments.

| ID | Sign | Scene Description | Category | Corresponding Traffic Rules |
|---|---|---|---|---|
| 1 | 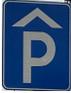 | Vehicle traveling on a curved road segment. | Parking | Parking is permitted only within the designated parking area indicated by the sign and must not obstruct traffic. |
| 2 | 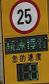 | Vehicle navigating a curving road segment. | Maximum Speed Limit 25 | Vehicles must not exceed a maximum speed of 25 km/h when driving on this road. |
| 3 | 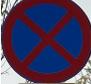 | Vehicle is driving through a curved portion of the road. | No Parking | Stopping or parking is strictly prohibited on this road sections. |
| 4 | 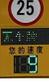 | A road user is traveling around a curved roadway segment. | Maximum Speed Limit 25 | Drivers must comply with the 25 km/h speed limit on this road segment. |
| 5 | 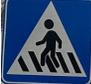 | Vehicle traveling on a curved road segment. | Pedestrian Crossing | Drivers must slow down and yield to pedestrians at the designated pedestrian crossing. |
| 6 | 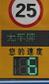 | Vehicle is driving through a curved portion of the road. | Maximum Speed Limit 25 | Vehicles must not exceed a maximum speed of 25 km/h when traveling on this road segment. |
| 7 | 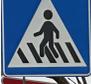 | Vehicle navigating a curving road segment. | Pedestrian Crossing | Drivers are required to yield to pedestrians crossing the road at the marked pedestrian crossing. |
| 8 | 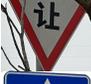 | Vehicle traveling on a right-angle road segment. | Yield | Drivers must yield the right of way to other vehicles or pedestrians when approaching the intersection. |
| 9 | 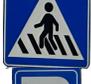 | Vehicle is approaching a right-angle turn section. | Pedestrian Crossing | Drivers must slow down and yield to pedestrians at the designated pedestrian crossing. |
| 10 | 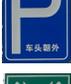 | Vehicle navigating a curving road segment. | Parking | Parking is permitted only within the designated parking area indicated by the sign and must not obstruct traffic. |
| 11 | 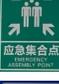 | Vehicle traveling on a straight road segment. | Emergency Station | Vehicles must not obstruct access to the emergency station and should yield to pedestrians and vehicles in the vicinity of the station. |
| 12 | 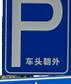 | The vehicle is traveling on a straight stretch of road. | Parking | Parking is permitted in the designated area indicated by the parking sign. |
| 13 | 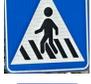 | A road user is driving along a linear road segment. | Pedestrian Crossing | Drivers must slow down and yield to pedestrians at the designated pedestrian crossing. |
| 14 | 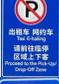 | The vehicle is traveling on a straight stretch of road. | No Parking | Stopping or parking is prohibited on this road segment. |
| 15 | 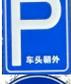 | Vehicle traveling on a straight road segment. | Parking | Parking is permitted in the designated area indicated by the parking sign. |
| 16 | 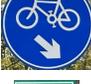 | Vehicle is advancing along a straight traffic lane. | Bicycle Lane | Motor vehicles are not permitted to enter or use the designated bicycle lane, which is reserved exclusively for bicycle traffic. |
| 17 | 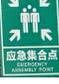 | The vehicle is traveling on a straight stretch of road. | Emergency Station | Stopping or parking is restricted in the vicinity of the emergency station to ensure unobstructed access for emergency vehicles and pedestrians. |

TABLE I: Examples of scene descriptions and corresponding rules (continued).

| ID | Sign | Scene Description | Category | Corresponding Traffic Rules |
|----|------|-------------------|----------|-----------------------------|
| 18 | 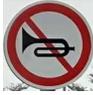 | Vehicle is advancing along a straight traffic lane. | No Honking | The use of vehicle horns is prohibited in this area except in situations where it is necessary to avoid immediate danger. |
| 19 | 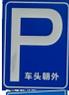 | Vehicle traveling on a cross intersection. | Parking | Parking is permitted at the intersection in accordance with the conditions indicated by the parking sign. |
| 20 | 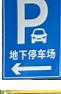 | Vehicle is entering a cross-shaped intersection area. | Parking | Parking is permitted at the intersection in accordance with the conditions specified by the parking sign. |
| 21 | 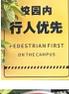 | Vehicle is approaching the center of a cross intersection. | Yield Here to Pedestrians | Vehicles must yield the right of way to pedestrians at the designated location within the intersection. |
| 22 | 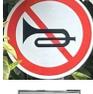 | The vehicle is driving across a crossroad junction. | No Honking | The use of vehicle horns is prohibited at the intersection except when necessary to prevent immediate danger. |
| 23 | 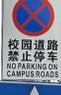 | Vehicle traveling on a cross intersection. | No Parking in Zone | Parking is prohibited within the designated zone at the intersection as indicated by the sign. |
| 24 | 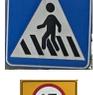 | The vehicle is driving across a crossroad junction. | Pedestrian Crossing | Vehicles are required to yield to pedestrians using the designated pedestrian crossing at the intersection. |
| 25 | 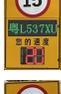 | Vehicle is entering a cross-shaped intersection area. | Maximum Speed Limit 15 | Vehicles must not exceed the maximum speed limit of 15 as indicated by the speed limit sign at the intersection. |
| 26 | 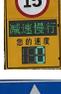 | Vehicle is approaching the center of a cross intersection. | Maximum Speed Limit 15 | Vehicles are required to comply with the maximum speed limit of 15 when traveling through the intersection. |
| 27 | 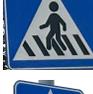 | Vehicle traveling on a curved road segment. | Pedestrian Crossing | Vehicles are required to yield to pedestrians using the designated pedestrian crossing on the curved road segment. |
| 28 | 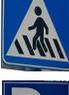 | A car is traveling through a curved section of roadway. | Pedestrian Crossing | Drivers must yield the right of way to pedestrians at the designated pedestrian crossing along the curved road. |
| 29 | 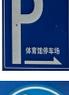 | The vehicle is moving through a curved road environment. | Parking | Parking is permitted on the curved road segment in accordance with the conditions indicated by the parking sign. |
| 30 | 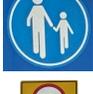 | Vehicle is negotiating a curved roadway section. | Sidewalk | Motor vehicles are not permitted to enter or use the sidewalk, which is reserved exclusively for pedestrian use. |
| 31 | 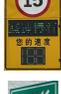 | Vehicle traveling on a curved road segment. | Maximum Speed Limit 15 | Vehicles must not exceed the maximum speed limit of 15 as indicated by the speed limit sign on the curved road segment. |
| 32 | 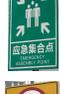 | The vehicle is moving through a curved road environment. | Emergency Station | Stopping or parking is restricted in the vicinity of the emergency station to ensure unobstructed access for emergency vehicles. |
| 33 | 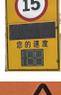 | A car is traveling through a curved section of roadway. | Maximum Speed Limit 15 | Vehicles are required to comply with the maximum speed limit of 15 when traveling along the curved road segment. |
| 34 | 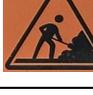 | The vehicle is moving through a curved road environment. | Road Work | Vehicles are required to proceed with caution and comply with temporary traffic regulations in areas affected by road work. |



\end{document}